\documentclass{article} % For LaTeX2e
\usepackage{iclr2021_conference,times}

\usepackage{amsmath,amsfonts,bm}

\def\eqref#1{equation~\ref{#1}}
\def\1{\bm{1}}

\DeclareMathAlphabet{\mathsfit}{\encodingdefault}{\sfdefault}{m}{sl}
\SetMathAlphabet{\mathsfit}{bold}{\encodingdefault}{\sfdefault}{bx}{n}

\usepackage{hyperref}

\usepackage{url}
\usepackage{float}

\usepackage{amssymb}

\usepackage{graphicx}
\usepackage[english]{babel}
\usepackage{comment}
\usepackage[T1]{fontenc}

\title{On Hard Episodes in meta-Learning}

\author{Samyadeep Basu, Amr Sharaf \\
Microsoft AI\\
\texttt{\{sbasu,amrsharaf\}@microsoft.com} \\
\And 
Nicolo Fusi \\
Microsoft Research \\
\texttt{fusi@microsoft.com} \\ 
\And
Soheil Feizi \\
Department of Computer Science \\
University of Maryland, College Park  \\
\texttt{\{sfeizi\}@cs.umd.edu} \\

}

\iclrfinalcopy % Uncomment for camera-ready version, but NOT for submission.
\begin{document}

\maketitle

\begin{abstract}

Existing meta-learners primarily focus on improving the average task accuracy across multiple episodes. Different
episodes, however, may vary in hardness and quality leading to a wide gap in the meta-learner's performance across
episodes. Understanding this issue is particularly critical in industrial few-shot settings, where there is limited
control over test episodes as they are typically uploaded by end-users. In this paper, we empirically analyse the
behaviour of meta-learners on episodes of varying hardness across three standard benchmark datasets: CIFAR-FS,
mini-ImageNet, and tiered-ImageNet. Surprisingly, we observe a wide gap in accuracy of around $50\%$ between the hardest
and easiest episodes across all the standard benchmarks and meta-learners. We additionally investigate various
properties of hard episodes and highlight their connection to catastrophic forgetting during meta-training.  To address
the issue of sub-par performance on hard episodes, we investigate and benchmark different meta-training strategies based
on adversarial training and curriculum learning.  We find that adversarial training strategies are much more powerful
than curriculum learning in improving the prediction performance on hard episodes.

%Existing meta-learners primarily focus on improving the average task accuracy across multiple episodes. However, different episodes may vary in hardness and quality leading to a wide gap in prediction performance across episodes. For example, in practical industrial few-shot settings, meta-learners trained on a base dataset are deployed for the end-users to upload their few-shot datasets (akin to episodes) in order to make predictions on new samples. These few-shot datasets often vary in quality and hardness as there is no control over the data that the end-users will upload. In our paper, we first  empirically analyse the behaviour of meta-learners on episodes of varying hardness across standard benchmark datasets such as CIFAR-FS, mini-ImageNet and tiered-ImageNet. To our surprise, we observe a wide gap in accuracy of around  $50\%$ between the hardest and easiest episodes across all the standard benchmarks and meta-learners. Second, we investigate various intriguing properties of hard episodes and find their connections with catastrophic forgetting during meta-training. To address the issue of subpar performance on hard episodes, we investigate and benchmark different meta-training strategies based on adversarial training and curriculum learning which moderately improves prediction performance on these hard episodes. 
\end{abstract}

\section{Introduction}

Humans have a remarkable ability to learn new concepts from very few examples and generalize effectively to unseen
tasks. However, standard deep learning approaches still lag behind human capabilities in learning from few examples. For
large over-parameterized deep models, learning with general supervision from only a few examples leads to over-fitting
and thus poor generalization. To circumvent this, the paradigm of few-shot learning~\citep{few_shot_survey,
Fei-fei06one-shotlearning, vinyals2017matching} aims to effectively learn new concepts from very few labeled examples.
These learned concepts can generalize well to future unseen learning tasks. Several frameworks have been proposed for
tackling the few-shot learning scenario: transfer-learning~\citep{few_shot_transfer},
self-training~\citep{self_training} and meta-learning~\citep{meta_learn_survey, maml_meta_learn, proto_meta_learn}.
Meta-learning in particular aims to learn the process of learning from few examples and has shown remarkable performance
across various few-shot benchmarks \citep{meta_learn_survey}.  In meta-learning, several few-shot tasks (episodes) are
sampled from a set of base classes and the underlying model is trained to perform well on these tasks leading to
improved generalization in learning from only few examples belonging to novel and unseen classes.

Existing meta-learners such as prototypical networks~\citep{proto_meta_learn}, MAML~\citep{maml_meta_learn},
MetaOptNet~\citep{meta_opt_net}, and R2D2~\citep{R2D2} primarily focus on improving prediction performance {\it on
average} across multiple episodes. However, different episodes have distinct characteristics and hardness which might
lead to a wide variance in prediction accuracy across episodes. This problem is much more prevalent in few-shot models
deployed in the industry. For example, meta-trained models are often deployed in the cloud for the end-users to use for
various tasks such as object recognition, detection, semantic segmentation in computer vision and natural language
understanding in NLP. In such settings, the end-users upload their own few-shot dataset to perform predictions on new
and unseen examples belonging to novel classes. In practice, different users may upload few-shot datasets of varying
quality and hardness, leading to a wide disparity in performance across different users.  To draw a parallel to the
widely accepted experimental protocols in meta-learning, each of the uploaded few-shot dataset and the corresponding
unseen examples is equivalent to a test episode. 

In this paper, we study this issue and investigate how existing state-of-the-art meta-learners~\citep{proto_meta_learn,
R2D2, meta_opt_net}  perform on episodes of varying hardness. Across three benchmark datasets: CIFAR-FS, mini-ImageNet, and
tieredImageNet, we observe that there is a gap of $\approx 50\%$ in prediction accuracy between the easiest and hardest
episodes. To this end, we identify several intriguing properties of hard episodes in meta-learning. For instance, we
find that hard episodes are {\it forgotten} more easily than easy episodes during meta-training. To improve prediction
performance on hard episodes, we investigate and benchmark various adversarial training and curriculum learning
strategies that can be used jointly with any existing meta-learner. Empirically, we find that adversarial training
strategies are much more powerful than curriculum learning in improving the prediction performance on hard episodes. The
aim of our paper is not to chase another state-of-the-art in meta-learning, but to perform a fine-grained inspection of
hard episodes across various meta-learning methods. 

In summary, we make the following contributions:
\begin{itemize}
    \item We present a detailed analysis of episode hardness in meta-learning across few-shot benchmarks and
	    state-of-the-art meta-learners. In particular, we study various properties (e.g., semantic characteristics,
	    forgetting) of episode hardness across different meta-learners and architectures.
    \item We find strong connections between episode hardness and catastrophic forgetting in meta-learning. While
	    catastrophic forgetting can occur when meta-training with multiple datasets in sequence
	    \citep{fewshot_forgetting}, we observe that forgetting events can occur even when the tasks during
	    meta-training are drawn from a single dataset. In particular, we find that hard episodes are easy to forget,
	    while easy episodes are difficult to forget.
    \item Based on our analysis, we investigate and benchmark different adversarial training and curriculum training
	    strategies to augment general purpose meta-training for improving prediction performance on hard episodes.
	    Empirically, we find that although there is no one-size-fits-all solution, adversarial meta-training
	    strategies are more powerful when compared to curriculum learning strategies. 
\end{itemize}

\section{Background and Related Work}

Meta-learning aims to learn an underlying model that can generalize and adapt well to examples from unseen classes by
the process of learning to learn. This is primarily achieved by mimicking the evaluation and adaptation procedure during
meta-training. In general, there are three types of meta-learners: (a) Memory-based
methods~\citep{Ravi2017OptimizationAA, munkhdalai2018rapid, memory_augmented} adapt to novel classes with a memory
attached to the meta-learner; (b) Metric-learning based methods~\citep{proto_meta_learn, relation_network} aim to learn
transferable deep representations which can adapt to unseen classes without any additional fine-tuning; (c) Optimization
based methods~\citep{maml_meta_learn, meta_opt_net, R2D2} learn a good pre-training initialization for effective
transfer to unseen tasks with only a few optimization steps. Although the primary focus of our work is meta-learning, we
note that other few-shot learning paradigms such as transfer learning \citep{chen2021metabaseline, sun2019metatransfer,
dhillon2020baseline} have also shown competitive performance with meta-learning.

While there has been a significant progress in improving the state-of-the-art in meta-learning, very few work
investigates the effectiveness of existing meta-learning approaches on episodes of varying hardness. A recent and
concurrent work by \cite{arnold2021uniform} discusses episode difficulty and the impact of random episodic sampling during
meta-training. Based on their analysis, \cite{arnold2021uniform} propose a re-weighted optimization framework for
meta-training based on importance sampling. Although our paper and \cite{arnold2021uniform} tackle similar problems of
episodic hardness, there are several points which distinguishes our work: 

\begin{itemize}
\item We provide a much more fine-grained analysis of episode hardness than \cite{arnold2021uniform}.
	\cite{arnold2021uniform} primarily discuss the transferability of episodes across different meta-learners, while
	we find and investigate a strong connection between episode hardness and catastrophic forgetting. 
\item \cite{arnold2021uniform} propose a loss re-weighting framework for improving the average accuracy across episodes.
	In contrary, we investigate the effectiveness of adversarial training \citep{maxup_gong} and general curriculum
	learning techniques in improving the average as well as worst-case prediction performance in meta-learning.  
\end{itemize}

Adversarial meta-learning techniques have previously been used in conjunction with data-augmentation
\citep{pmlr-v139-ni21a} to select the augmentation type resulting in the worst-case loss among different augmentation
techniques. In this paper, we focus on how such strategies can be useful in improving the prediction performance of the
hard episodes in addition to the average accuracy.

\begin{figure*}
    \hskip-0.8cm
  \includegraphics[width=15.3cm, height=5.0cm]{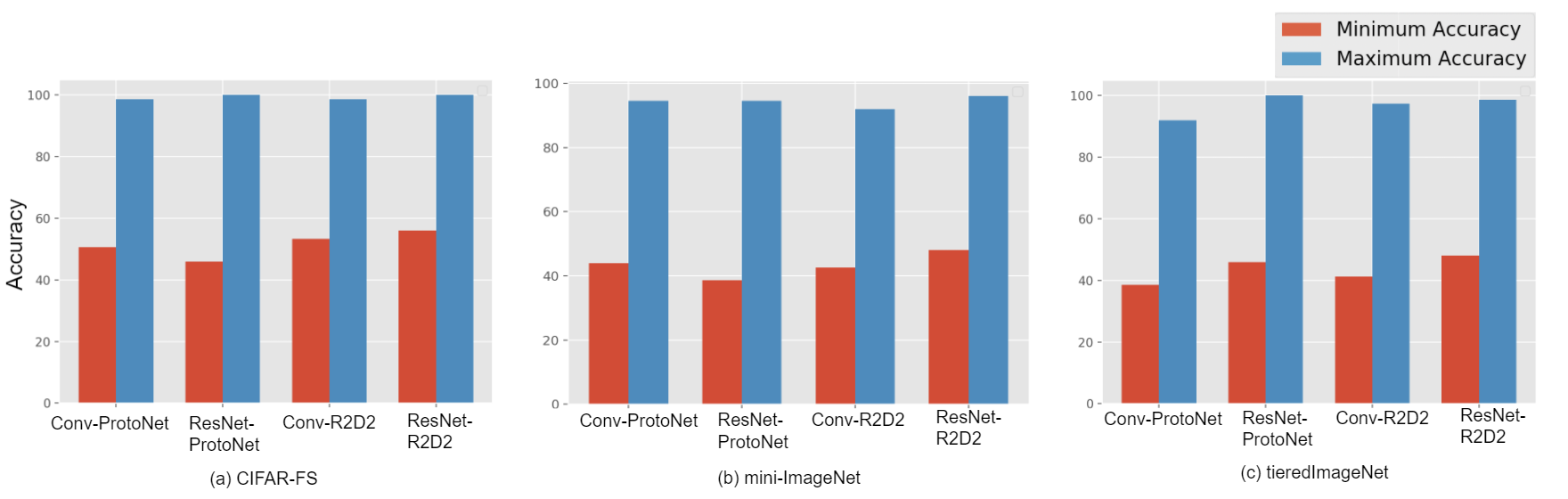}
  \vspace{-2em}
  \caption{\label{fig_hardness} Accuracy (y-axis) of existing meta-learners on the hardest and the easiest episode across standard few-shot datasets and meta-learners (x-axis). Note that there is a wide gap of $\approx 50 \%$ between the prediction performance on the easiest and hardest episode.}
\end{figure*}

\section{Rethinking Episodic Accuracy}

Existing state-of-the-art meta-learners~\citep{maml_meta_learn, meta_opt_net, proto_meta_learn, R2D2} primarily focus on
optimizing for the average loss across multiple training episodes or tasks. However, solely the average performance in
isolation does not give enough insights into how meta-learners perform on episodes of varying quality and hardness. Such
insights can be particularly crucial to investigate and debug meta-learning models deployed in the wild, where the model
can encounter diverse test episodes. In this section, we go beyond the average accuracy across different test episodes
and evaluate meta-learners on episodes of varying hardness.  First, we discuss how to quantify the hardness of an
episode and then discuss the performance of meta-learners on hard episodes. 
\vspace{-1em}
\subsection{What is a good measure of episode hardness?}

Episodic sampling (i.e. sampling various few-shot tasks from a base dataset) in meta-learning takes place in two steps:
(i) First the episode classes are sampled from the class distribution of the base classes : $ c \sim
p(\mathcal{C}_{base})$; (ii) Next, an episode $\tau$ is sampled from the data distribution conditioned on the set of
sampled classes $c$: $\tau \sim p(\mathcal{D}|c)$, where $\mathcal{D}$ is the base dataset. An episode $\tau$ consists
of a set of support examples $\tau_{s}$ and query examples $\tau_{q}$. In few-shot learning, a $n$-way, $k$-shot episode
is sampled which results in sampling $n$ classes and $k$ support examples per class. Based on this, the meta-learning
optimization objective can be generalized as the following:
\begin{equation}
    \theta^{*} =\arg \min_{\theta} \mathbb{E}_{\tau}[\ell(\mathcal{F}_{\theta^{'}}, \tau_{q})]
\end{equation}
where $\mathcal{F}$ is the base architecture with $\theta$ as the model parameters and $\theta^{'} = \mathcal{A}(\theta,
\tau_{s})$ is the fine-tuning step with the support examples. Different meta-learners have different types of
fine-tuning procedures and we direct the readers to \citep{maml_meta_learn, proto_meta_learn, R2D2} for more information
on the characteristics of $\mathcal{A}$. Based on this definition, we define the hardness of an episode
$\mathcal{H}(\tau)$ in terms of the loss incurred on the query examples in an episode:
\begin{equation}
    \mathcal{H}(\tau) = \ell(\mathcal{F}_{\theta^{*}}, \tau_{q})
\end{equation}
% TODO add forward pointers to appendix
We choose query loss as a metric for hardness because of its inherent simplicity in computation as well as
interpretation. In addition, we find a strong negative correlation between the episodic loss and the accuracy ($\approx
-0.92$ for mini-ImageNet and $\approx -0.89$ for tieredImageNet with prototypical networks). This is true for other
meta-learners such as R2D2 too (See Appendix A for more details). Alternatively, hardness of an episode can also be defined as the average log-odds of the query example \citep{dhillon2020baseline}.
\vspace{-1em}
\subsection{Performance of meta-learners on hard episodes}
To understand the effectiveness of meta-learners on episodes of varying hardness, we first order the test episodes in
decreasing order of their hardness. Then, we evaluate different meta-learners on the easiest and the hardest test
episode. Across all the few-shot benchmark datasets such as mini-ImageNet, CIFAR-FS and tieredImageNet, we find in Fig.
(\ref{fig_hardness}) that there is a gap in accuracy of $\approx 50 \%$ between the episodes with the highest and the
lowest loss. Furthermore, we find that the meta-learner and architecture which performs well on average, does not
necessarily perform well on the hardest episode. For example, in the case of mini-ImageNet, prototypical networks with a
stronger architecture such as ResNet performs better than the 4-layered convolutional architecture on an average, but
not on the hard episodes. Moreover in Fig. (\ref{fig_hardness}), we notice that for tieredImageNet, prototypical
networks with ResNet performs the best on easy episodes, while R2D2 with ResNet performs slightly better on the hard
episodes. The wide gap in accuracy between hard and easy episodes can be magnified for meta-learners deployed in the
wild, where the model might encounter episodes which are significantly hard and diverse in nature. Going forward, we
believe that meta-learning studies should not only report average episodic accuracy, but also the prediction accuracy on
easy and hard episodes, to present a complete picture of the effectiveness of the meta-learner. 

%In the next sections, we study various properties of episodes which incur high loss.

% \subsection{Note On Transferability of episodes}
% Concurrent to our work, \citep{arnold2021uniform} has shown that the hardness of episodes transfer across different meta-learners and base architectures. We validate their findings and extend their analysis to gain new insights into the transferability of episodes (See Appendix for more details).  In particular, we find a distinction between the transferability of training episodes and test episodes. Notably, we observe that the transfer of training episodes is significantly more noisy than the transfer of test episodes. Amongst the training episodes, we notice that episodes which are hard, transfer much more effectively than their easy counterparts.

\section{Visual Semantics of Hard Episodes}

Based on the observed disparity in performance between hard and easy episodes, a natural question arises: what causes
certain episodes to incur high loss? While quantitatively, the hardness of an episode can be defined in terms of the
loss incurred on the query examples, it does not provide salient insights into the qualitative nature of hard episodes.
In general, we find that episodes incur high loss when there is a mismatch in the semantic characteristics between the
support and query examples. For example, when a majority of the support examples have objects of only one category in
the frame and the query examples have multiple objects of different categories surrounding the primary object of
interest, the underlying meta-learner often leads to a wrong prediction. Furthermore, when the shape of the objects in
the query examples is slightly different from the objects in the support examples, the prediction is often erroneous. 

%\red{use either For example or just e.g.,},

\begin{figure}[H]
    \hskip-0.5cm
  \includegraphics[width=15.3cm, height=3.1cm]{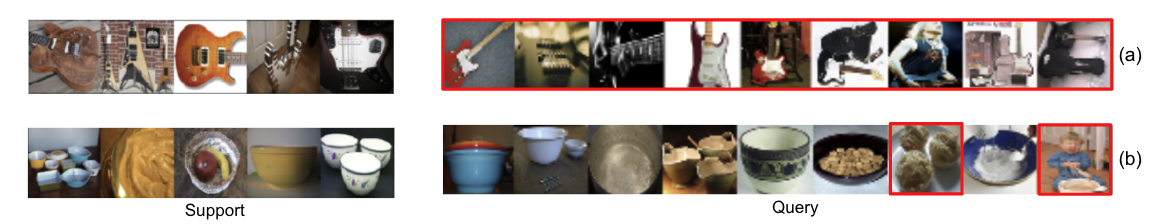}
    \vspace{-2em}
  \caption{\label{semantic} Semantic Properties of Hard and Easy Episodes: (a) Hard episode, Class: electric-guitar; (b) Easy  episode, Class: mixing-bowl; The images marked in red borders are misclassified query examples.}
\end{figure}

In Fig. (\ref{semantic})-(a), we notice that the query examples that have different objects (e.g., humans) along with
the primary object (i.e., guitar) are often misclassified. In Fig. (\ref{semantic})-(b), where most of the query
examples are classified correctly, we find that the misclassified examples are of two types: (i) query images in which
the primary object is occluded with a secondary object; (ii) the shape of the object in the query example is different
from the object shapes in the support examples. We provide additional examples of hard episodes in Appendix E.

%In the next section, we find and discuss a strong connection between episode hardness and catastrophic forgetting in meta-learning. 
% \begin{figure*}
%     \hskip-0.10cm
%   \includegraphics[width=14.9cm, height=3.3cm]{hard_eps.png}
%   \vspace{-2em}
%   \caption{\label{hard_ep}Semantic characteristics of an hard episode.(to change with high res)}
% \end{figure*}
% \begin{figure*}
%     \hskip-0.10cm
%   \includegraphics[width=14.7cm, height=3.3cm]{easy_eps.png}
%   \vspace{-2em}
%   \caption{\label{easy_ep}Semantic characteristics of an easy episode. (to change with high res)}
% \end{figure*}

\section{Hard Episodes Suffer from Forgetting } 
\label{hard_forget}
\vspace{-0.5em}
In supervised learning, catastrophic forgetting is prevalent when tasks from different distributions are learned
sequentially~\citep{catastrophic_forget}. In such cases, old tasks are forgotten as the model encounters and learns from
new tasks. \cite{toneva2019empirical} has shown that certain examples can be forgotten with high frequency during the
course of supervised training even when the samples are drawn from a single dataset. In meta-learning,
\cite{fewshot_forgetting} has shown that in meta-training with tasks from {\it multiple} task distributions
sequentially, tasks from the old distribution in the sequence can be forgotten as the meta-learner encounters new tasks.
However, we observe that even in the case of meta-training with tasks drawn from a {\it single} task distribution,
certain types of tasks (episodes) can be forgotten during the course of training. In particular, we analyze the
connection between the hardness of episodes and catastrophic forgetting in meta-learning. We track the behaviour of easy
and hard episodes during meta-training and in summary find that:

% . In this section, we investigate catastrophic forgetting in the context of meta-learning where the tasks are drawn only from a single dataset \red{it is repetitive, combine these two paragraphs}. In particular, we track the behaviour of certain episodes during the course of meta-training and in summary find that: \red{In these two paragraphs, we never mention hard episodes. the connection to the previous section sounds loose.}

(i) For hard episodes, we notice that the final accuracy at the end of the training drops significantly ($\approx 15 \%$ in
some cases) from the maximum accuracy obtained during the course of meta-training. 

(ii) Hard episodes have more number of forgetting events in comparison to the easy episodes during the course of
meta-training.  This behaviour is more pronounced in the later stages of meta-training where the accuracies of the easy
episodes have already stabilized.
\begin{figure*}
    \hskip-1cm
  \includegraphics[width=15.8cm, height=4.5cm]{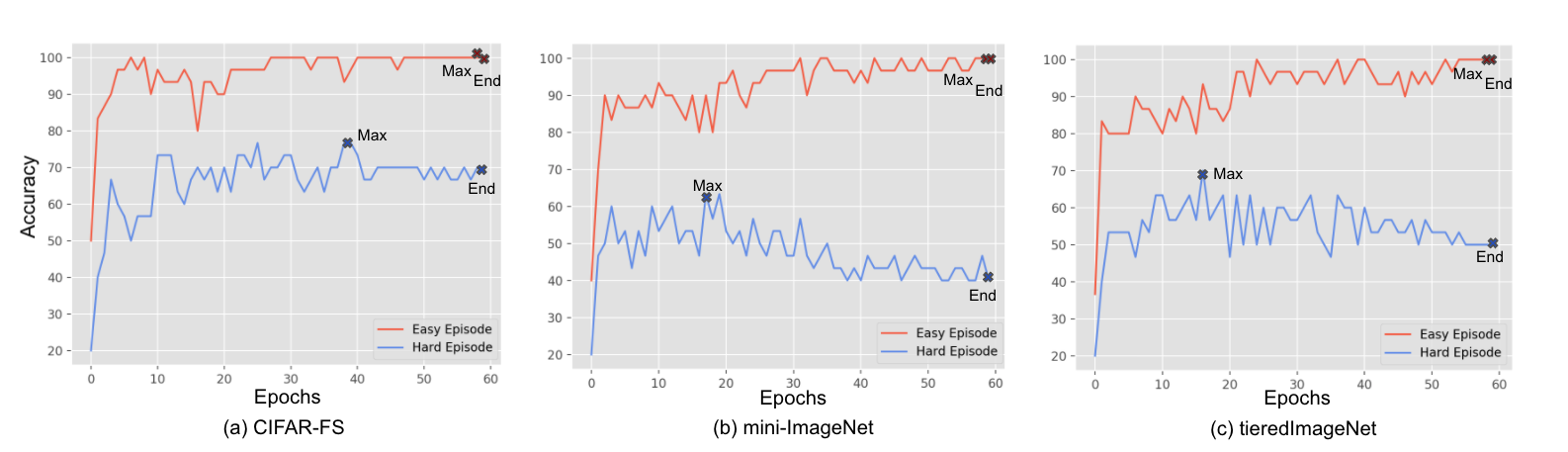}
  \vspace{-2em}
  \caption{\label{local_forgetting}Accuracy of easy and hard episodes (y-axis) during the course of meta-training across different epochs (x-axis); Hard episodes often have a final accuracy less than the maximum accuracy reached during meta-training. }
\end{figure*}
%\red{It would have been nice to somehow highlight the max and end accuracies for the plots, or highlight the gap between them using a two sided arrow} 
\subsection{Defining Forgetting Events in meta learning}

During the course of meta-training, the set of sampled tasks are different in each epoch. In order to track forgetting
events during meta-training, we first randomly select a set of $k$ episodes ($\mathcal{E}= \{\tau\}_{i=1}^{k}$) and
track their accuracy, throughout the course of meta-training. In our experiments, we set $k=160$ . We primarily define
two types of forgetting events: (i) Local forgetting event; (ii) Global forgetting event.

\textbf{Global forgetting events}. For a given episode, a global forgetting event is encountered if the accuracy of the
episode at the end of meta-training is less than the maximum accuracy reached during the course of training by a particular threshold. Formally,
given an episode $\tau$ with the maximum accuracy $acc_{max}(\tau) = \max_{j} acc_{j}(\tau)$, a global forgetting event
occurs if $acc_{max}(\tau) \geq acc_{end}(\tau) + \alpha$, where $\alpha$ is a threshold and $acc_{end}(\tau)$ is the
accuracy at the end of meta-training. Note that for each episode, a global forgetting event can occur only once. 

\textbf{Local forgetting events}. For an episode $\tau$ in the $j^{th}$ epoch of meta-training, a local forgetting event is
encountered if the accuracy of the episode at the $j^{th}$ epoch ($acc_{j}(\tau)$) is less than the accuracy at the $(j-1)^{th}$
epoch ($acc_{j-1}(\tau)$) by a particular threshold, denoted by $\alpha$. Formally, a local forgotten event is encountered if $acc_{j}(\tau) + \alpha
\leq acc_{j-1}(\tau) $. Empirically, we study local forgetting events for $ 0.03 \leq \alpha \leq 0.15$.

\subsection{Forgetting Events and Hard Episodes}

\begin{figure}[H]
    \hskip-0.95cm
  \includegraphics[width=15.6cm, height=4.8cm]{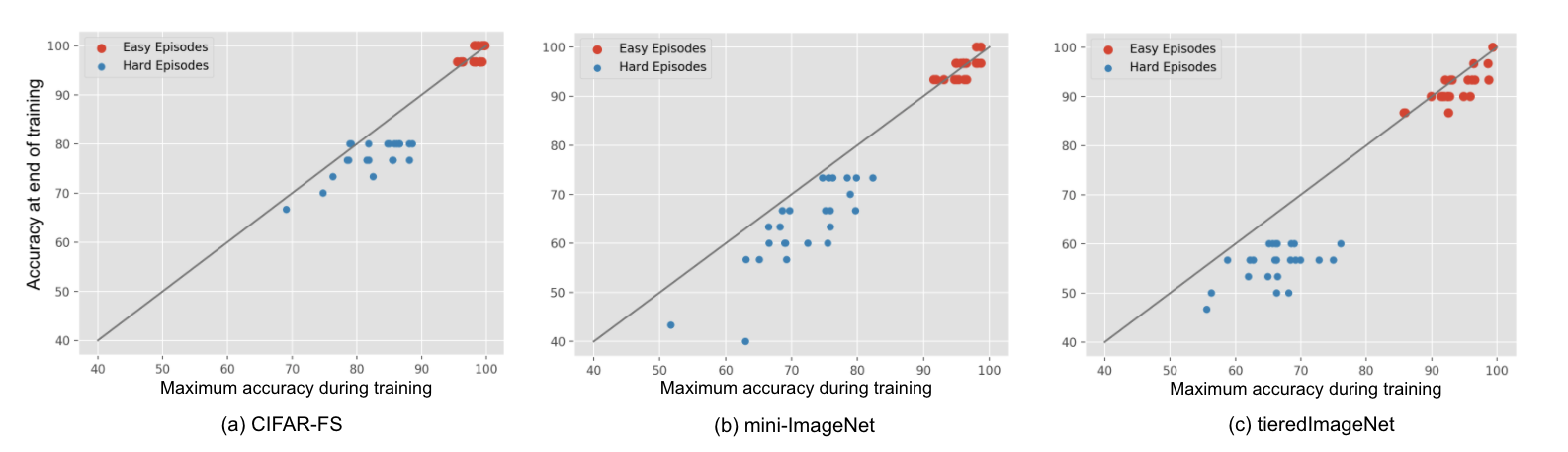}
    \vspace{-2em}
  \caption{\label{global} Hard episodes have a wider gap between the final accuracy (y-axis) and the maximum accuracy reached during meta-training (x-axis), in comparison to easy episodes. This behaviour is more pronounced for mini-ImageNet and tieredImageNet.}
\end{figure}

\textbf{Global forgetting events.} In Fig. (\ref{local_forgetting}), we track the accuracy of the hardest and easiest
episode from each of the few-shot datasets during the entire course of meta-training across different epochs. Visually,
we observe that for the hard episode, the accuracy decreases after a certain point during the course of meta-training.
However for the easy episode, this is not the case and the accuracy increases till the end of meta-training.  To draw
more insights,  we compute the global forgetting behaviour of different episodes. We first choose the 15 hardest and
easiest episodes respectively from CIFAR-FS, mini-ImageNet, tieredImageNet and compute their final episodic accuracy and
the maximum episodic accuracy reached during meta-training. In Fig. (\ref{global}), we find that for the hard episodes,
the gap between the final accuracy and the maximum accuracy reached during meta-training is significantly larger than
the easy episodes. The gap in particular is large for mini-ImageNet and tieredImageNet, while for CIFAR-FS the gap is
relatively narrow. For example, in case of the mini-ImageNet dataset, the gap can be $\approx 15 \%$, whereas for the
tieredImageNet dataset, this gap can be $\approx 10 \%$. For CIFAR-FS, this gap is $\approx 6 \%$.  This characteristic
shows that hard episodes are globally forgotten in comparison to the easier episodes during the entire course of
meta-training. We provide further results on global forgetting events in Appendix C.

\begin{figure*}
    \hskip-0.8cm
  \includegraphics[width=15.6cm, height=4.7cm]{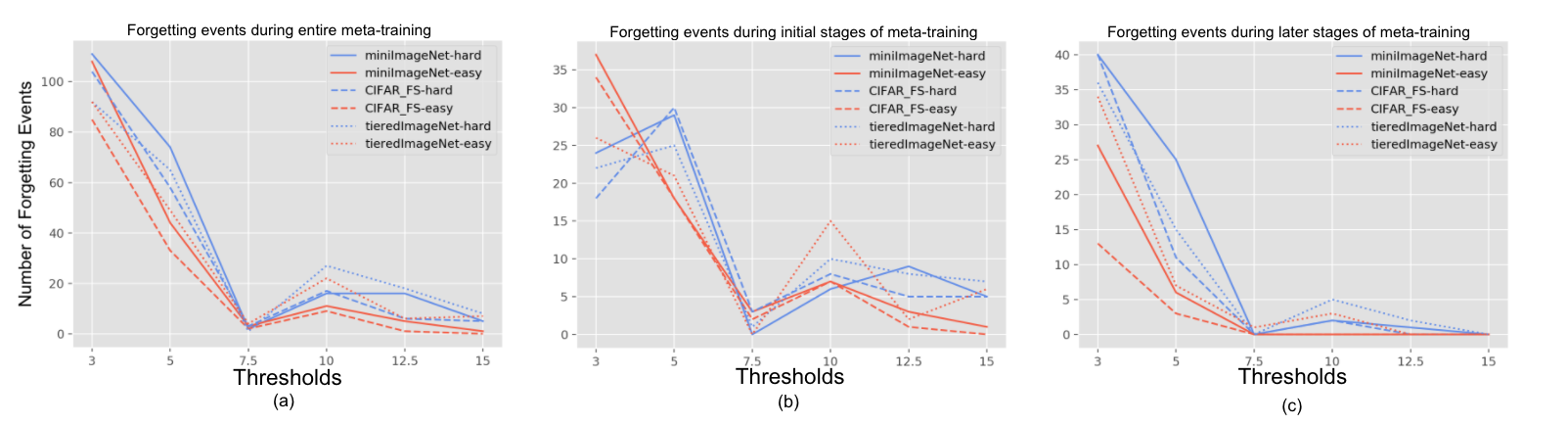}
  \vspace{-2em}
  \caption{\label{global_events} (a) Total number of local forgetting events (y-axis) across different thresholds
  (x-axis) during the course of meta-training; (b) Total number of local forgetting events during the first 20 epochs of
  meta-training; (c) Total number of local forgetting events during the last 20 epochs of meta-training. The number of
  local forgetting events is higher for hard episodes in comparison to the easy episodes across different thresholds.}
\end{figure*}

\textbf{Local forgetting events.} In order to compute the frequency of local forgetting events for easy and hard
episodes across the three few-shot datasets, we first choose 15 easy and hard episodes from each dataset. Across this
set of easy and hard episodes, we then compute the number of local forgetting events across various thresholds. In
general, across the entire duration of meta-training, we find that the hard episodes have more local forgetting events
than the easy episodes. For instance, in Fig. (\ref{global_events})-(a), we observe that there is a substantial gap in
the number of forgetting events encountered for easy and hard episodes during the entire course of meta-training.
Furthermore, to gain more insights about this gap in the number of encountered forgetting events, we understand how this
gap behaves in the first 20 epochs of meta-training (Fig. (\ref{global_events})-(b)) and the last 20 epochs of
meta-training (Fig. (\ref{global_events})-(c)). In particular, we find that the gap is narrow during the initial stages
of meta-training, whereas the gap widens substantially during the later stages. 

To summarize, we find that forgetting occurs in meta-learning even when the tasks are drawn from a { \it single} task
distribution. Furthermore, we find a strong connection between episode hardness and forgetting events, where we show
that hard episodes are more easily forgotten than easy episodes, both in the local and global contexts. In the next
section, we investigate two meta-training strategies to improve few-shot performance on hard episodes. 
\section{Improving Performance on Hard Episodes}

In this section, we investigate and benchmark two different meta-training strategies based on adversarial training and
curriculum learning in order to improve prediction performance on hard episodes. Recent work \citep{maxup_gong,
pmlr-v139-ni21a} uses adversarial training to select episode specific augmentations from a wide pool of diverse data
augmentation methods resulting in the highest loss. This loss is then optimized with respect to the model parameters
during training. Such data augmentation selection strategies have been shown to mitigate over-fitting and improve
generalization. In our work, we study how such training strategies can be used to select hard episodes and improve
prediction performance across these episodes. In particular, we investigate two variants of adversarial training as
proposed first in \cite{maxup_gong}: (i) General adversarial training (AT); (ii) Adversarial curriculum training (ACT).
\vspace{-0.5em}
\subsection{General adversarial training}
We adopt the adversarial training procedure from \citep{maxup_gong} to first select episodes with a high loss and
optimize the meta-learner only with respect to the loss incurred by such hard episodes. Specifically, this involves
solving a saddle-point optimization problem where the underlying loss is minimized with respect to the parameters of the
model and maximized with respect to the input. In particular, during each update of the meta-learner, we first draw a
batch of episodes each containing support and query examples. Then for each element of the batch, we additionally sample
a number of similar episodes and select the episode with the highest loss. The model parameters are then updated with
respect to the gradient of the loss incurred by the selected hard episode. Formally, we solve the following min-max
optimization: 
\begin{equation}
\label{adv_train}
\min_{\theta} \mathbb{E}_{\tau}[ \max_{t \in g(\mathcal{\tau})} \ell(\mathcal{F}_{\theta^{'}}, t_{q})]
\end{equation}
where $\theta^{'} = \mathcal{A}(\theta, t_{s})$ is the fine-tuning step with the support examples from the selected
episodes with the highest loss, $g(\tau)$ is an operator which samples additional episodes to select from for each task
$\tau$ in the batch, $\mathcal{F}$ is the base learner and $\ell$ is the loss function. In our experiments, we let
$g(\tau)$ select four additional episodes per sampled episode $\tau$. We provide a more detailed description of the
hyper-parameters in Appendix B.1. 
\subsection{Adversarial Curriculum Training}
Curriculum learning \citep{curriculum_1, curriculum_2} aims to mimic the learning process of humans and animals. In
particular, humans learn new tasks in a well defined order; i.e., it first learns easy tasks and then gradually moves
towards learning more complex tasks. Inspired by this, we modify Eq. (\ref{adv_train}) and investigate a curriculum
meta-training strategy. Specifically, during the initial phase of meta-training, we select only a set of easy episodes
to learn from, while in the later stages of training, harder episodes are selected. Formally, given the underlying model
is meta-trained for $|e|$ epochs, for the first $|e|/2$ epochs, the following loss is optimized:
\begin{equation}
\label{adv_curr_train_easy}
\min_{\theta} \mathbb{E}_{\tau}[ \min_{t \in g(\mathcal{\tau})} \ell(\mathcal{F}_{\theta^{'}}, t_{q})]
\end{equation}
and for the last $|e|/2$ epochs during meta-training, the general min-max adversarial loss described in Eq. (\ref{adv_train}) is used. 
\section{Experiments}
\subsection{Experimental Setup}
\textbf{Datasets}. We use three standard few-shot classification datasets for our experiments : (i) CIFAR-FS
\citep{R2D2}; (ii) mini-ImageNet \citep{mini_imagenet} and (iii) tieredImageNet \citep{tiered_imagenet}. CIFAR-FS is
sampled from CIFAR-100, whereas mini-ImageNet and tieredImageNet are subsets of ImageNet \citep{imagenet1k}. We use the
class splits from \citep{meta_opt_net} for all the three datasets. Note that tieredImageNet is a more challenging
dataset than mini-ImageNet as the splits are constructed from near the root of the ImageNet hierarchy. We provide more
details on the datasets in Appendix B.2.
\begin{table}
\hspace{3.5em}
\begin{tabular}{|l|l|l|l|l|l|l|}
\hline
\textbf{Method} & \multicolumn{2}{l|}{\textbf{CIFAR-FS}} & \multicolumn{2}{l|}{\textbf{mini-ImageNet}} & \multicolumn{2}{l|}{\textbf{tieredImageNet}} \\ \hline
 & 1-shot & 5-shot & 1-shot & 5-shot & 1-shot & 5-shot \\ \hline
\begin{tabular}[c]{@{}l@{}}Conv-ProtoNet\\ + AT\\ + ACT\end{tabular} & \begin{tabular}[c]{@{}l@{}}62.6\\ \textbf{63.3}\\ 63.1\end{tabular} & \begin{tabular}[c]{@{}l@{}}80.9\\ 80.8\\ 80.0\end{tabular} & \begin{tabular}[c]{@{}l@{}}52.3\\ \textbf{52.7}\\ 52.4\end{tabular} & \begin{tabular}[c]{@{}l@{}}70.4\\ \textbf{70.8}\\ 70.3\end{tabular} & \begin{tabular}[c]{@{}l@{}}52.6\\ 52.7\\ \textbf{52.9}\end{tabular} & \begin{tabular}[c]{@{}l@{}}71.5\\ \textbf{71.9}\\ 71.5\end{tabular} \\ \hline
\begin{tabular}[c]{@{}l@{}}ResNet12-ProtoNet\\ + AT\\ + ACT\end{tabular} & \begin{tabular}[c]{@{}l@{}}71.0\\ \textbf{71.2}\\ 69.3\end{tabular} & \begin{tabular}[c]{@{}l@{}}84.0\\ 83.9\\ 82.6\end{tabular} & \begin{tabular}[c]{@{}l@{}}59.0\\ 58.9\\ 56.0\end{tabular} & \begin{tabular}[c]{@{}l@{}}74.8\\ \textbf{74.9}\\ 72.8\end{tabular} & \begin{tabular}[c]{@{}l@{}}61.8\\ \textbf{62.9}\\ \textbf{63.4}\end{tabular} & \begin{tabular}[c]{@{}l@{}}80.0\\ \textbf{80.4}\\ 79.9\end{tabular} \\ \hline
\begin{tabular}[c]{@{}l@{}}Conv-R2D2\\ +AT\\ +ACT\end{tabular} & \begin{tabular}[c]{@{}l@{}}67.6\\ \textbf{68.0}\\ \textbf{68.3}\end{tabular} & \begin{tabular}[c]{@{}l@{}}82.6\\ \textbf{82.8}\\ 82.6\end{tabular} & \begin{tabular}[c]{@{}l@{}}55.3\\ \textbf{55.7}\\ 55.4\end{tabular} & \begin{tabular}[c]{@{}l@{}}72.4\\ 72.4\\ 71.3\end{tabular} & \begin{tabular}[c]{@{}l@{}}56.8\\ \textbf{57.2}\\ \textbf{57.8}\end{tabular} & \begin{tabular}[c]{@{}l@{}}75.00\\ 75.00\\ 74.9\end{tabular} \\ \hline
\begin{tabular}[c]{@{}l@{}}ResNet12-R2D2\\ +AT\\ +ACT\end{tabular} & \begin{tabular}[c]{@{}l@{}}70.0\\ \textbf{71.0}\\ 69.4\end{tabular} & \begin{tabular}[c]{@{}l@{}}84.8\\ 84.9\\ 83.2\end{tabular} & \begin{tabular}[c]{@{}l@{}}58.6\\ 58.63\\ 56.3\end{tabular} & \begin{tabular}[c]{@{}l@{}}75.5\\ \textbf{76.0}\\ 74.4\end{tabular} & \begin{tabular}[c]{@{}l@{}}62.8\\ \textbf{63.7}\\ \textbf{63.5}\end{tabular} & \begin{tabular}[c]{@{}l@{}}80.4\\ \textbf{81.3}\\ 80.9\end{tabular} \\ \hline
\begin{tabular}[c]{@{}l@{}}ResNet12 -MetaOptNet\\ +AT\\ +ACT\end{tabular} & \begin{tabular}[c]{@{}l@{}}70.8\\ \textbf{71.2}\\ 70.2\end{tabular} & \begin{tabular}[c]{@{}l@{}}84.0\\ \textbf{84.6}\\ 83.9\end{tabular} & \begin{tabular}[c]{@{}l@{}}60.1\\ \textbf{60.9}\\ 59.8\end{tabular} & \begin{tabular}[c]{@{}l@{}}77.4\\ \textbf{78.2}\\ 77.1\end{tabular} &  \begin{tabular}[c]{@{}l@{}}62.9\\ \textbf{63.3}\\ 62.1\end{tabular} & \begin{tabular}[c]{@{}l@{}}80.7\\ 80.9\\ 79.8\end{tabular} \\ \hline
\end{tabular}
\caption{\label{avg_perform}Average episodic performance of general adversarial training (AT) and adversarial curriculum training (ACT) across different meta-learners.}
\end{table}

\textbf{Architectures}. We primarily use two standard architectures for our experiments: (i) 4-layer convolutional network as introduced in \citep{vinyals2017matching}; (ii) ResNet-12 \citep{resnet} which is used by \citep{tadam} in the few-shot setting. Both these architectures use batch-normalization after every convolutional layer and use ReLU as the activation function. Similar architectures for few-shot learning have been previously used in \citep{meta_opt_net, arnold2021uniform}.

\textbf{Meta-learners}. We use prototypical networks \citep{proto_meta_learn} from the metric learning family of few-shot algorithms. In addition, we use MetaOptNet \citep{meta_opt_net} and R2D2 \citep{R2D2} as representative algorithms from the optimization based meta-learning methods. 

\textbf{Model Selection}. We use the validation set for each dataset to select the best model. Primarily, we run the validation procedure every 1k iterations on 2k episodes from the validation set to select the best model. Finally, we evaluate on 1k test episodes from the test set for each dataset. 
\vspace{-0.9em}
\subsection{Discussion}
\begin{table}
\hspace{3.5em}
\begin{tabular}{|l|l|l|l|l|l|l|}
\hline
\textbf{Method} & \multicolumn{2}{l|}{\textbf{CIFAR-FS}} & \multicolumn{2}{l|}{\textbf{mini-ImageNet}} & \multicolumn{2}{l|}{\textbf{tieredImageNet}} \\ \hline
 & 1-shot & 5-shot & 1-shot & 5-shot & 1-shot & 5-shot \\ \hline
\begin{tabular}[c]{@{}l@{}}Conv-ProtoNet\\ + AT\\ + ACT\end{tabular} & \begin{tabular}[c]{@{}l@{}}28.8\\ \textbf{29.3}\\ 28.5\end{tabular} & \begin{tabular}[c]{@{}l@{}}63.0\\ 62.9\\ 61.3\end{tabular} & \begin{tabular}[c]{@{}l@{}}20.2\\ 19.3\\ 19.4\end{tabular} & \begin{tabular}[c]{@{}l@{}}53.2\\ \textbf{53.5}\\ 52.7\end{tabular} & \begin{tabular}[c]{@{}l@{}}22.2\\ \textbf{24.6}\\ 22.1\end{tabular} & \begin{tabular}[c]{@{}l@{}}52.3\\ \textbf{52.7}\\ \textbf{53.1}\end{tabular} \\ \hline
\begin{tabular}[c]{@{}l@{}}ResNet12-ProtoNet\\ + AT\\ + ACT\end{tabular} & \begin{tabular}[c]{@{}l@{}}36.0\\ 33.7\\ 29.46\end{tabular} & \begin{tabular}[c]{@{}l@{}}66.7\\ 66.7\\ 64.9\end{tabular} & \begin{tabular}[c]{@{}l@{}}26.2\\ \textbf{29.2}\\ 24.5\end{tabular} & \begin{tabular}[c]{@{}l@{}}56.5\\ \textbf{58.3}\\ 54.7\end{tabular} & \begin{tabular}[c]{@{}l@{}}29.8\\ \textbf{31.2}\\ 29.6\end{tabular} & \begin{tabular}[c]{@{}l@{}}60.4\\ \textbf{60.9}\\ 60.4\end{tabular} \\ \hline
\begin{tabular}[c]{@{}l@{}}Conv-R2D2\\ +AT\\ +ACT\end{tabular} & \begin{tabular}[c]{@{}l@{}}32.2\\ \textbf{35.3}\\ 32.8\end{tabular} & \begin{tabular}[c]{@{}l@{}}65.3\\ \textbf{65.7}\\ 65.5\end{tabular} & \begin{tabular}[c]{@{}l@{}}21.4\\ \textbf{24.0}\\ \textbf{23.7}\end{tabular} & \begin{tabular}[c]{@{}l@{}}55.7\\ 55.6\\ 54.5\end{tabular} & \begin{tabular}[c]{@{}l@{}}27.0\\ \textbf{27.6}\\ \textbf{27.8}\end{tabular} & \begin{tabular}[c]{@{}l@{}}56.6\\ 56.6\\ 56.7\end{tabular} \\ \hline
\begin{tabular}[c]{@{}l@{}}ResNet12-R2D2\\ +AT\\ +ACT\end{tabular} & \begin{tabular}[c]{@{}l@{}}30.6\\ \textbf{35.6}\\ 31.8\end{tabular} & \begin{tabular}[c]{@{}l@{}}68.0\\ 68.0\\ 65.9\end{tabular} & \begin{tabular}[c]{@{}l@{}}28.0\\ \textbf{28.5}\\ 24.4\end{tabular} & \begin{tabular}[c]{@{}l@{}}59.0\\ \textbf{60.1}\\ 56.8\end{tabular} & \begin{tabular}[c]{@{}l@{}}34.0\\ 33.8\\ 30.8\end{tabular} & \begin{tabular}[c]{@{}l@{}}61.5\\ \textbf{62.1}\\ \textbf{62.0}\end{tabular} \\ \hline
\begin{tabular}[c]{@{}l@{}}ResNet12 -MetaOptNet\\ +AT\\ +ACT\end{tabular} & \begin{tabular}[c]{@{}l@{}}37.2\\ \textbf{37.9}\\ 37.1\end{tabular} & \begin{tabular}[c]{@{}l@{}}69.2\\ \textbf{70.0}\\ 69.1\end{tabular} & \begin{tabular}[c]{@{}l@{}}29.5\\ \textbf{31.2}\\ 29.2\end{tabular} & \begin{tabular}[c]{@{}l@{}}61.1\\ \textbf{62.3}\\ 60.5\end{tabular} & \begin{tabular}[c]{@{}l@{}}30.4\\ \textbf{31.5}\\ 30.2\end{tabular}  &  \begin{tabular}[c]{@{}l@{}}61.5\\ 61.4\\ 60.9\end{tabular} \\ \hline
\end{tabular}
\caption{\label{hard_results} Performance of general adversarial training (AT) and adversarial curriculum training (ACT) across different meta-learners on {\it hard episodes.} We report the mean accuracy over 30 hardest episodes for each meta-learner.}
\end{table}
\subsubsection{Adversarial training improves performance on hard episodes}
Across different meta-learners (ProtoNets, R2D2, MetaOptNet) and few-shot datasets (CIFAR-FS, mini-ImageNet, tieredImageNet), we find that the adversarial training (AT) strategy works well in general to improve both the average episodic performance as well as the episodic performance on hard episodes. The comprehensive results for the average performance of meta-learners is presented in Table (\ref{avg_perform}), while the results on hard episodes are presented in Table (\ref{hard_results}). In particular, we find that adversarial meta-training strategies never hurt the average episodic performance and improves over the baseline in a majority of our experimental settings. However, we find that the adversarial training strategy (AT) leads to a large gain over the baseline meta-training strategy for hard episodes. Specifically, we find that the improvements are more significant for the 1-shot case when compared to the 5-shot case. For example. specific to the 1-shot case, we observe a $5\%$ gain for CIFAR-FS with R2-D2 and $3\%$ gain for mini-ImageNet with prototypical networks. For tieredImageNet, we observe $\approx 2\%$ improvement on episodic performance for hard episodes with prototypical networks.
\subsubsection{Adversarial training is better than curriculum training}
Although curriculum training leads to better generalization in supervised learning \citep{curriculum_1,curriculum_2}, we find that in meta-learning, the adversarial curriculum strategy (ACT) generally performs worse than both the baseline and general adversarial training (AT) in a majority of our experimental settings. Our observation on curriculum training for meta-learning is consistent with the recent work of \citep{arnold2021uniform} where they show that curriculum meta-training strategies underperform significantly when compared to the baseline meta-training. We note that although the curriculum formulation in \citep{arnold2021uniform} is different than ours, both methods present easy episodes to the meta-learner first followed by hard episodes. While we present a negative result on curriculum meta-training, we believe that this observation can be used as a note to develop more advanced and improved curriculum meta-training strategies in the future. 

In summary, we find that although there is no one-size-fits-all solution to improve performance on hard episodes, adversarial meta-training strategies perform  better than the baseline and curriculum learning.
\vspace{-0.5em}
\section{Conclusion}
\vspace{-0.5em}
In our paper, we investigated how different meta-learners perform on episodes of varying hardness. We found that there exists a wide gap in the performance of the meta-learners between the easiest and the hardest episode across different few-shot datasets. Furthermore, we investigated various facets of the hard episodes and uncovered two major properties: (i) Hard episodes usually have multiple diverse objects in the query examples, whereas the support set primarily consists of objects of a single category, (ii) Hard episodes are forgotten by the underlying meta-learner at a higher frequency than the easier episodes. To improve prediction performance over hard episodes, we investigated and benchmarked different training strategies such as adversarial training and curriculum learning during meta-training. We found that adversarial training strategies are beneficial towards meta-training and improve the average episodic performance as well as performance over hard episodes. Based on our analysis in this paper, designing more robust meta-learning algorithms which can generalize to hard episodes is an important direction of future work.

% \red{Based on our analysis in this paper, we believe that understanding how existing few-shot learners perform on test-episodes with different kinds of distributional shifts is an important direction of future work.}\red{SF: maybe instead of this say that the average accuracy may not be sufficient to fully characterize the performance of few-shot learners and we need ... } 
\section{Reproducibility Statement}
Small experimental details are crucial to reproduce results in deep learning. In order to foster the reproducibility of our paper, we provide all the small details of our experiments in the Appendix (Section : Hyperparameters). We would like to point out that we use the general hyperparameters in our experimental setup, as used in \citep{meta_opt_net} to ensure reproducibility of the baselines which is particularly important in few-shot learning research. Additionally, considering a majority of our paper is on analysis of meta-learners and generating plots, we provide all the necessary details (e.g, number of episodes chosen for generating the plots) available in each corresponding section (See the subsections of Section \ref{hard_forget}). 

\section{Ethics Statement}
In recent times, there has been a significant increase in the usage of deep learning models in the industry. Domains of application include NLP, computer vision and healthcare to name a few. Training task and domain specific deep models require large amounts of labeled training data which is not available for a variety of use-cases. The paradigm of few-shot learning (or meta-learning) aims to learn generalizable models from only a few training examples. Our paper is primarily geared towards understanding the science of such few-shot learning methods. We believe our work can accelerate research on few-shot learning with a specific focus on improving learning from hard episodes. To the best of our knowledge our work will facilitate few-shot learning research and does not lead to any negative or unethical societal impact.

% \section{Modified Elastic Weight Consolidation for Meta-Learning}
% One way to improve upon hard episodes is to reduce forgetting during the meta-training stage. To this regard, we use elastic weight consolidation from supervised learning and adapt it to meta-learning. Using elastic weight consolidation for meta-learning is not trivial and we modify the elastic weight consolidation regularizer. The modified loss function with the regularizer at the $t^{th}$ epoch of meta-training:
% \begin{equation}
%     \min_{\theta} \{\frac{1}{|B|} \sum_{i=1}^{|B|} \ell(\tau_{i}, \theta)  \} + \lambda \sum_{j}^{n} g(F_{j,j}^{t}, F_{j,j}^{t-1}) (\theta - \theta_{j}^{t-1})^{2}
% \end{equation}
% where:
% \[
%   g(F_{j,j}^{t}, F_{j,j}^{t-1}) = 
% \begin{cases}
%      0 ,& \text{if } sgn(\frac{1}{|B|} \sum_{k=1}^{|B|} \frac{\partial \ell(\tau_{k}^{t})}{\partial \theta_{j} }) = sgn(\frac{1}{|B|} \sum_{k=1}^{|B|} \frac{\partial \ell(\tau_{k}^{t-1})}{\partial \theta_{j}^{t-1} })\\
%     F_{j,j}^{t-1},              & \text{otherwise}
% \end{cases}
% \]
% and:
% \begin{equation}
%     F_{j,j}^{t} = \frac{1}{|B|}\sum_{k=1}^{|B|} (\frac{\partial \ell(\tau_{k}^{t})}{\partial \theta_{j} })^{2}
% \end{equation}

% \section{Elastic Weight Consolidation with Meta-Training}
% \begin{equation}
%     \ell^{t}(\theta_{F}) = \ell^{t}(\theta) + \lambda \sum_{i}^{} F_{i,i} (\theta_{i} - \theta^{*}_{t-\alpha, i})^{2}
% \end{equation}
% where:
% \begin{equation}
%     F_{i,i} = \frac{1}{|B|}\sum_{j=1}^{|B|} (\frac{\partial  (\ell^{t-\alpha}(\tau_{j}, \theta))}{\partial \theta_{i}})^{2}
% \end{equation}

\bibliography{iclr2021_conference}

\begin{thebibliography}{32}
\providecommand{\natexlab}[1]{#1}
\providecommand{\url}[1]{\texttt{#1}}
\expandafter\ifx\csname urlstyle\endcsname\relax
  \providecommand{\doi}[1]{doi: #1}\else
  \providecommand{\doi}{doi: \begingroup \urlstyle{rm}\Url}\fi

\bibitem[Arnold et~al.(2021)Arnold, Dhillon, Ravichandran, and
  Soatto]{arnold2021uniform}
Sébastien M.~R. Arnold, Guneet~S. Dhillon, Avinash Ravichandran, and Stefano
  Soatto.
\newblock Uniform sampling over episode difficulty, 2021.

\bibitem[Bengio et~al.(2009)Bengio, Louradour, Collobert, and
  Weston]{curriculum_2}
Yoshua Bengio, J\'{e}r\^{o}me Louradour, Ronan Collobert, and Jason Weston.
\newblock Curriculum learning.
\newblock In \emph{Proceedings of the 26th Annual International Conference on
  Machine Learning}, ICML '09, pp.\  41–48, New York, NY, USA, 2009.
  Association for Computing Machinery.
\newblock ISBN 9781605585161.
\newblock \doi{10.1145/1553374.1553380}.
\newblock URL \url{https://doi.org/10.1145/1553374.1553380}.

\bibitem[Bertinetto et~al.(2018)Bertinetto, Henriques, Torr, and Vedaldi]{R2D2}
Luca Bertinetto, Jo{\~{a}}o~F. Henriques, Philip H.~S. Torr, and Andrea
  Vedaldi.
\newblock Meta-learning with differentiable closed-form solvers.
\newblock \emph{CoRR}, abs/1805.08136, 2018.
\newblock URL \url{http://arxiv.org/abs/1805.08136}.

\bibitem[Chen et~al.(2021)Chen, Liu, Xu, Darrell, and
  Wang]{chen2021metabaseline}
Yinbo Chen, Zhuang Liu, Huijuan Xu, Trevor Darrell, and Xiaolong Wang.
\newblock Meta-baseline: Exploring simple meta-learning for few-shot learning,
  2021.

\bibitem[Deng et~al.(2009)Deng, Dong, Socher, Li, Li, and Fei-Fei]{imagenet1k}
Jia Deng, Wei Dong, Richard Socher, Li-Jia Li, Kai Li, and Li~Fei-Fei.
\newblock Imagenet: A large-scale hierarchical image database.
\newblock In \emph{2009 IEEE Conference on Computer Vision and Pattern
  Recognition}, pp.\  248--255, 2009.
\newblock \doi{10.1109/CVPR.2009.5206848}.

\bibitem[Dhillon et~al.(2019)Dhillon, Chaudhari, Ravichandran, and
  Soatto]{few_shot_transfer}
Guneet~S. Dhillon, Pratik Chaudhari, Avinash Ravichandran, and Stefano Soatto.
\newblock A baseline for few-shot image classification.
\newblock \emph{CoRR}, abs/1909.02729, 2019.
\newblock URL \url{http://arxiv.org/abs/1909.02729}.

\bibitem[Dhillon et~al.(2020)Dhillon, Chaudhari, Ravichandran, and
  Soatto]{dhillon2020baseline}
Guneet~S. Dhillon, Pratik Chaudhari, Avinash Ravichandran, and Stefano Soatto.
\newblock A baseline for few-shot image classification, 2020.

\bibitem[Fei-fei et~al.(2006)Fei-fei, Fergus, and
  Perona]{Fei-fei06one-shotlearning}
Li~Fei-fei, Rob Fergus, and Pietro Perona.
\newblock One-shot learning of object categories.
\newblock \emph{IEEE TRANSACTIONS ON PATTERN ANALYSIS AND MACHINE
  INTELLIGENCE}, 28:\penalty0 2006, 2006.

\bibitem[Finn et~al.(2017)Finn, Abbeel, and Levine]{maml_meta_learn}
Chelsea Finn, Pieter Abbeel, and Sergey Levine.
\newblock Model-agnostic meta-learning for fast adaptation of deep networks.
\newblock \emph{CoRR}, abs/1703.03400, 2017.
\newblock URL \url{http://arxiv.org/abs/1703.03400}.

\bibitem[Gidaris \& Komodakis(2018)Gidaris and
  Komodakis]{dynamic_without_forgetting}
Spyros Gidaris and Nikos Komodakis.
\newblock Dynamic few-shot visual learning without forgetting.
\newblock \emph{CoRR}, abs/1804.09458, 2018.
\newblock URL \url{http://arxiv.org/abs/1804.09458}.

\bibitem[Gong et~al.(2020)Gong, Ren, Ye, and Liu]{maxup_gong}
ChengYue Gong, Tongzheng Ren, Mao Ye, and Qiang Liu.
\newblock Maxup: {A} simple way to improve generalization of neural network
  training.
\newblock \emph{CoRR}, abs/2002.09024, 2020.
\newblock URL \url{https://arxiv.org/abs/2002.09024}.

\bibitem[Hacohen \& Weinshall(2019)Hacohen and Weinshall]{curriculum_1}
Guy Hacohen and Daphna Weinshall.
\newblock On the power of curriculum learning in training deep networks.
\newblock \emph{CoRR}, abs/1904.03626, 2019.
\newblock URL \url{http://arxiv.org/abs/1904.03626}.

\bibitem[He et~al.(2015)He, Zhang, Ren, and Sun]{resnet}
Kaiming He, Xiangyu Zhang, Shaoqing Ren, and Jian Sun.
\newblock Deep residual learning for image recognition.
\newblock \emph{CoRR}, abs/1512.03385, 2015.
\newblock URL \url{http://arxiv.org/abs/1512.03385}.

\bibitem[Hospedales et~al.(2020)Hospedales, Antoniou, Micaelli, and
  Storkey]{meta_learn_survey}
Timothy~M. Hospedales, Antreas Antoniou, Paul Micaelli, and Amos~J. Storkey.
\newblock Meta-learning in neural networks: {A} survey.
\newblock \emph{CoRR}, abs/2004.05439, 2020.
\newblock URL \url{https://arxiv.org/abs/2004.05439}.

\bibitem[Kirkpatrick et~al.(2016)Kirkpatrick, Pascanu, Rabinowitz, Veness,
  Desjardins, Rusu, Milan, Quan, Ramalho, Grabska{-}Barwinska, Hassabis,
  Clopath, Kumaran, and Hadsell]{catastrophic_forget}
James Kirkpatrick, Razvan Pascanu, Neil~C. Rabinowitz, Joel Veness, Guillaume
  Desjardins, Andrei~A. Rusu, Kieran Milan, John Quan, Tiago Ramalho, Agnieszka
  Grabska{-}Barwinska, Demis Hassabis, Claudia Clopath, Dharshan Kumaran, and
  Raia Hadsell.
\newblock Overcoming catastrophic forgetting in neural networks.
\newblock \emph{CoRR}, abs/1612.00796, 2016.
\newblock URL \url{http://arxiv.org/abs/1612.00796}.

\bibitem[Lee et~al.(2019)Lee, Maji, Ravichandran, and Soatto]{meta_opt_net}
Kwonjoon Lee, Subhransu Maji, Avinash Ravichandran, and Stefano Soatto.
\newblock Meta-learning with differentiable convex optimization.
\newblock \emph{CoRR}, abs/1904.03758, 2019.
\newblock URL \url{http://arxiv.org/abs/1904.03758}.

\bibitem[Munkhdalai et~al.(2018)Munkhdalai, Yuan, Mehri, and
  Trischler]{munkhdalai2018rapid}
Tsendsuren Munkhdalai, Xingdi Yuan, Soroush Mehri, and Adam Trischler.
\newblock Rapid adaptation with conditionally shifted neurons, 2018.

\bibitem[Ni et~al.(2021)Ni, Goldblum, Sharaf, Kong, and
  Goldstein]{pmlr-v139-ni21a}
Renkun Ni, Micah Goldblum, Amr Sharaf, Kezhi Kong, and Tom Goldstein.
\newblock Data augmentation for meta-learning.
\newblock In Marina Meila and Tong Zhang (eds.), \emph{Proceedings of the 38th
  International Conference on Machine Learning}, volume 139 of
  \emph{Proceedings of Machine Learning Research}, pp.\  8152--8161. PMLR,
  18--24 Jul 2021.
\newblock URL \url{https://proceedings.mlr.press/v139/ni21a.html}.

\bibitem[Oreshkin et~al.(2018)Oreshkin, L{\'{o}}pez, and Lacoste]{tadam}
Boris~N. Oreshkin, Pau~Rodr{\'{\i}}guez L{\'{o}}pez, and Alexandre Lacoste.
\newblock {TADAM:} task dependent adaptive metric for improved few-shot
  learning.
\newblock \emph{CoRR}, abs/1805.10123, 2018.
\newblock URL \url{http://arxiv.org/abs/1805.10123}.

\bibitem[Phoo \& Hariharan(2020)Phoo and Hariharan]{self_training}
Cheng~Perng Phoo and Bharath Hariharan.
\newblock Self-training for few-shot transfer across extreme task differences.
\newblock \emph{CoRR}, abs/2010.07734, 2020.
\newblock URL \url{https://arxiv.org/abs/2010.07734}.

\bibitem[Qiao et~al.(2017)Qiao, Liu, Shen, and Yuille]{predict_param}
Siyuan Qiao, Chenxi Liu, Wei Shen, and Alan~L. Yuille.
\newblock Few-shot image recognition by predicting parameters from activations.
\newblock \emph{CoRR}, abs/1706.03466, 2017.
\newblock URL \url{http://arxiv.org/abs/1706.03466}.

\bibitem[Ravi \& Larochelle(2017)Ravi and Larochelle]{Ravi2017OptimizationAA}
S.~Ravi and H.~Larochelle.
\newblock Optimization as a model for few-shot learning.
\newblock In \emph{ICLR}, 2017.

\bibitem[Ren et~al.(2018)Ren, Triantafillou, Ravi, Snell, Swersky, Tenenbaum,
  Larochelle, and Zemel]{tiered_imagenet}
Mengye Ren, Eleni Triantafillou, Sachin Ravi, Jake Snell, Kevin Swersky,
  Joshua~B. Tenenbaum, Hugo Larochelle, and Richard~S. Zemel.
\newblock Meta-learning for semi-supervised few-shot classification.
\newblock \emph{CoRR}, abs/1803.00676, 2018.
\newblock URL \url{http://arxiv.org/abs/1803.00676}.

\bibitem[Santoro et~al.(2016)Santoro, Bartunov, Botvinick, Wierstra, and
  Lillicrap]{memory_augmented}
Adam Santoro, Sergey Bartunov, Matthew Botvinick, Daan Wierstra, and Timothy
  Lillicrap.
\newblock Meta-learning with memory-augmented neural networks.
\newblock In \emph{Proceedings of the 33rd International Conference on
  International Conference on Machine Learning - Volume 48}, ICML'16, pp.\
  1842–1850. JMLR.org, 2016.

\bibitem[Snell et~al.(2017)Snell, Swersky, and Zemel]{proto_meta_learn}
Jake Snell, Kevin Swersky, and Richard~S. Zemel.
\newblock Prototypical networks for few-shot learning.
\newblock \emph{CoRR}, abs/1703.05175, 2017.
\newblock URL \url{http://arxiv.org/abs/1703.05175}.

\bibitem[Sun et~al.(2019)Sun, Liu, Chua, and Schiele]{sun2019metatransfer}
Qianru Sun, Yaoyao Liu, Tat-Seng Chua, and Bernt Schiele.
\newblock Meta-transfer learning for few-shot learning, 2019.

\bibitem[Sung et~al.(2017)Sung, Yang, Zhang, Xiang, Torr, and
  Hospedales]{relation_network}
Flood Sung, Yongxin Yang, Li~Zhang, Tao Xiang, Philip H.~S. Torr, and
  Timothy~M. Hospedales.
\newblock Learning to compare: Relation network for few-shot learning.
\newblock \emph{CoRR}, abs/1711.06025, 2017.
\newblock URL \url{http://arxiv.org/abs/1711.06025}.

\bibitem[Toneva et~al.(2019)Toneva, Sordoni, des Combes, Trischler, Bengio, and
  Gordon]{toneva2019empirical}
Mariya Toneva, Alessandro Sordoni, Remi~Tachet des Combes, Adam Trischler,
  Yoshua Bengio, and Geoffrey~J. Gordon.
\newblock An empirical study of example forgetting during deep neural network
  learning, 2019.

\bibitem[Vinyals et~al.(2016)Vinyals, Blundell, Lillicrap, Kavukcuoglu, and
  Wierstra]{mini_imagenet}
Oriol Vinyals, Charles Blundell, Timothy~P. Lillicrap, Koray Kavukcuoglu, and
  Daan Wierstra.
\newblock Matching networks for one shot learning.
\newblock \emph{CoRR}, abs/1606.04080, 2016.
\newblock URL \url{http://arxiv.org/abs/1606.04080}.

\bibitem[Vinyals et~al.(2017)Vinyals, Blundell, Lillicrap, Kavukcuoglu, and
  Wierstra]{vinyals2017matching}
Oriol Vinyals, Charles Blundell, Timothy Lillicrap, Koray Kavukcuoglu, and Daan
  Wierstra.
\newblock Matching networks for one shot learning, 2017.

\bibitem[Wang et~al.(2020)Wang, Yao, Kwok, and Ni]{few_shot_survey}
Yaqing Wang, Quanming Yao, James~T. Kwok, and Lionel~M. Ni.
\newblock Generalizing from a few examples: A survey on few-shot learning.
\newblock \emph{ACM Comput. Surv.}, 53\penalty0 (3), June 2020.
\newblock ISSN 0360-0300.
\newblock \doi{10.1145/3386252}.
\newblock URL \url{https://doi.org/10.1145/3386252}.

\bibitem[Yap et~al.(2020)Yap, Ritter, and Barber]{fewshot_forgetting}
Pau~Ching Yap, Hippolyt Ritter, and David Barber.
\newblock Bayesian online meta-learning with laplace approximation.
\newblock \emph{CoRR}, abs/2005.00146, 2020.
\newblock URL \url{https://arxiv.org/abs/2005.00146}.

\end{thebibliography}
\bibliographystyle{iclr2021_conference}
\newpage
\appendix
\section{Loss as a measure of hardness}
\begin{figure}[H]
    \hskip0cm
  \includegraphics[width=14.0cm, height=10cm]{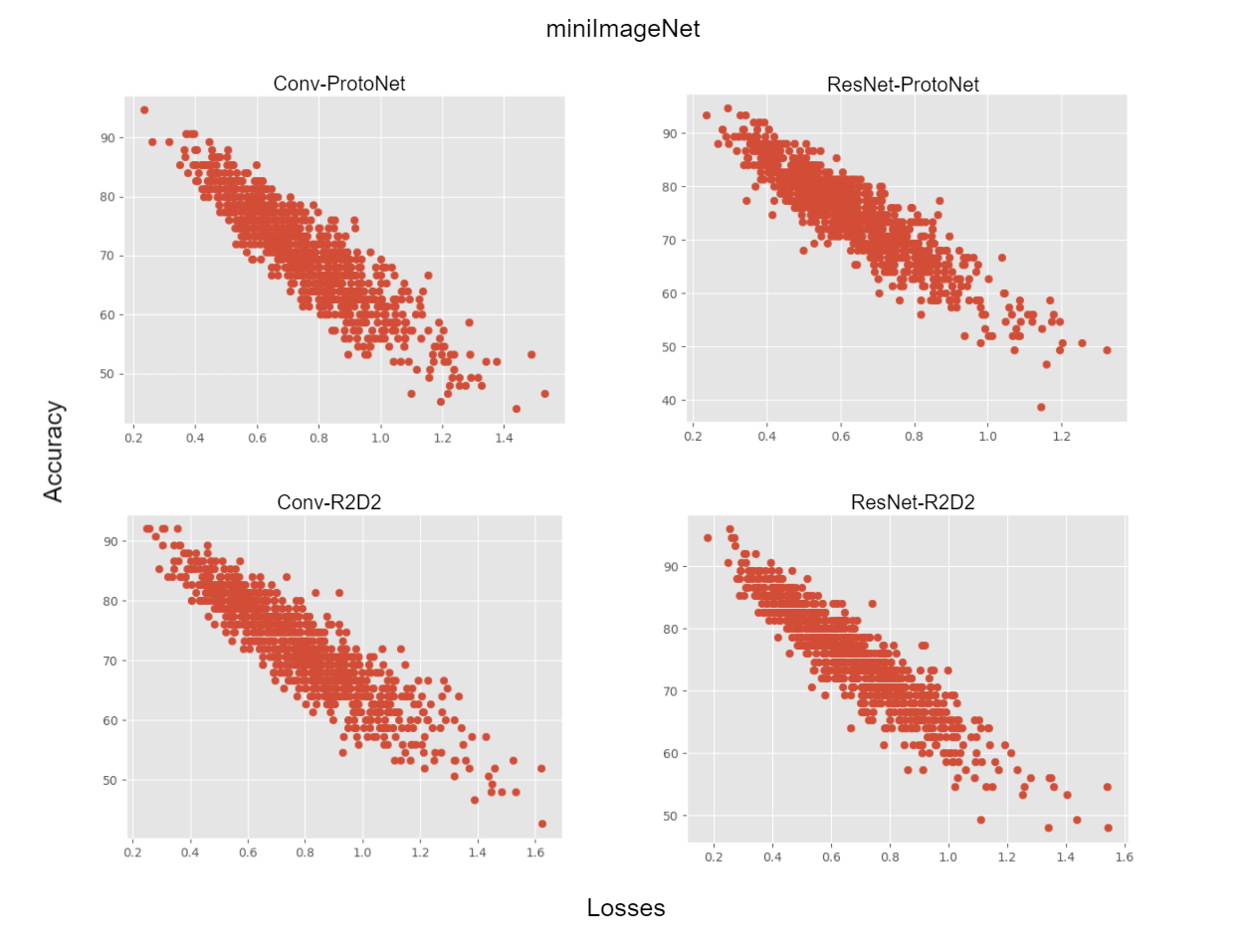}
  \vspace{-2em}
  \caption{\label{mini_losses} Loss (x-axis) vs. Accuracy (y-axis) plot for miniImageNet.}
\end{figure}

\begin{figure}[H]
    \hskip 0cm
  \includegraphics[width=14.0cm, height=10cm]{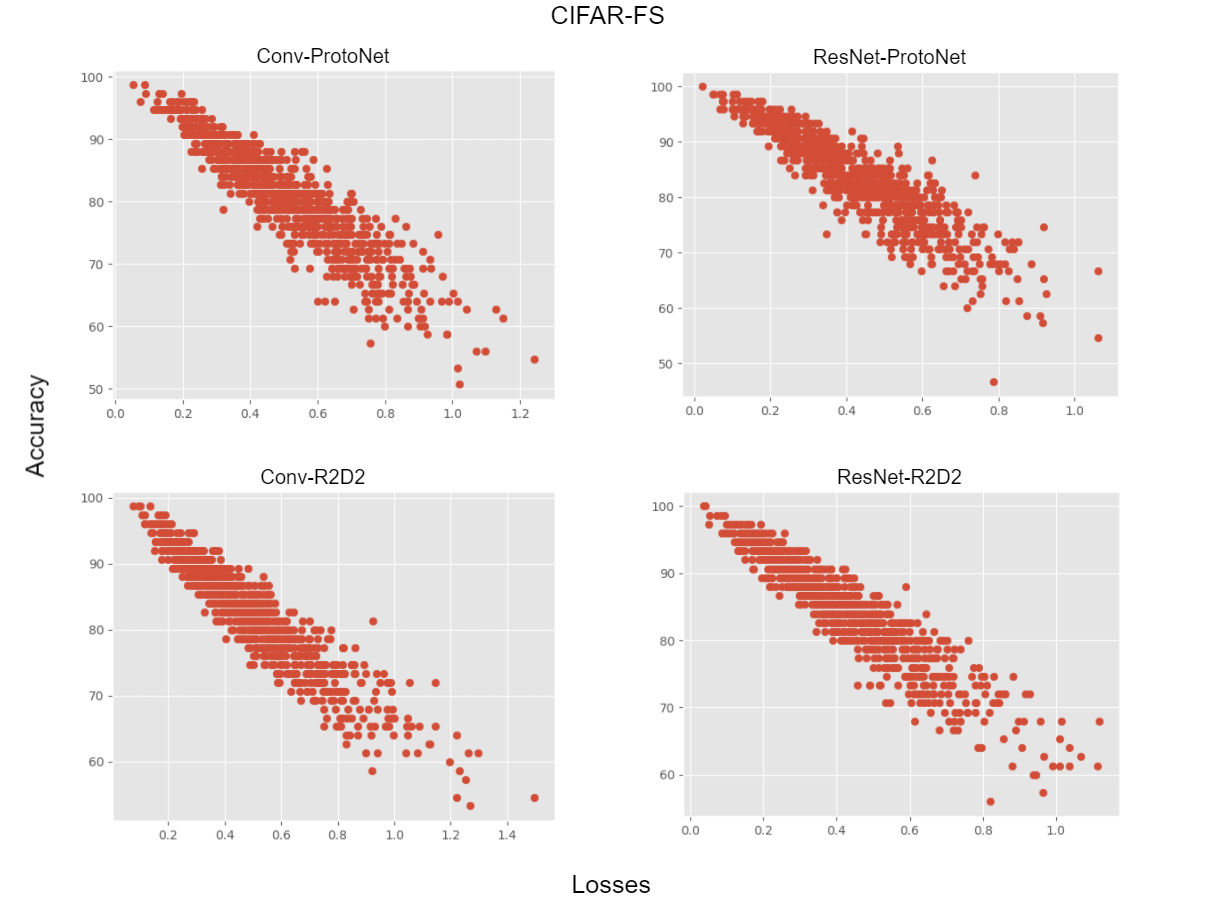}
  \vspace{-2em}
  \caption{\label{cifar_losses} Loss (x-axis) vs. Accuracy (y-axis) plot for CIFAR-FS.}
\end{figure}

\begin{figure}[H]
    \hskip 0cm
  \includegraphics[width=14.0cm, height=10.5cm]{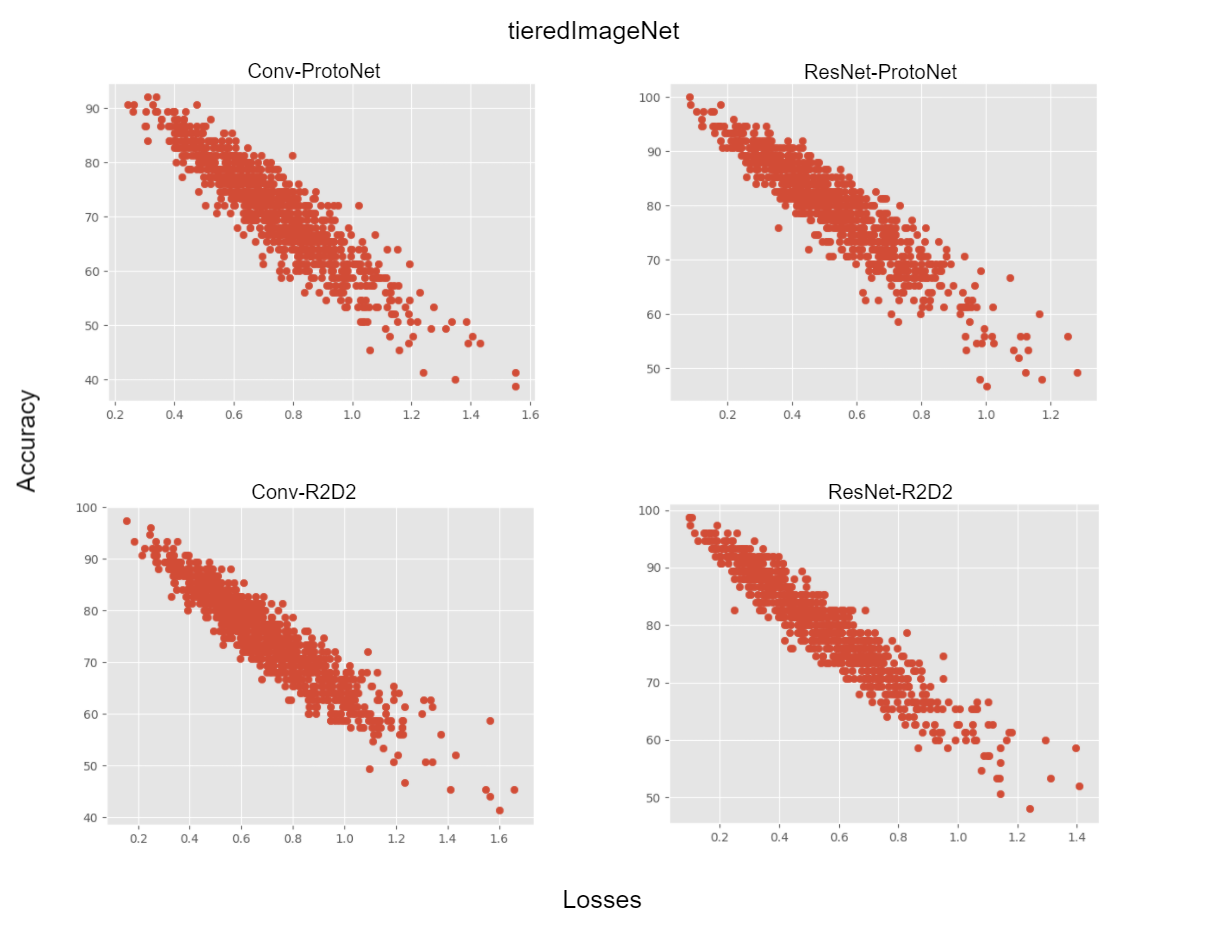}
  \vspace{-2em}
  \caption{\label{tiered_losses} Loss (x-axis) vs. Accuracy (y-axis) plot for tieredImageNet.}
\end{figure}
\begin{figure}[H]
    \hskip 1.5cm
  \includegraphics[width=11.0cm, height=8.5cm]{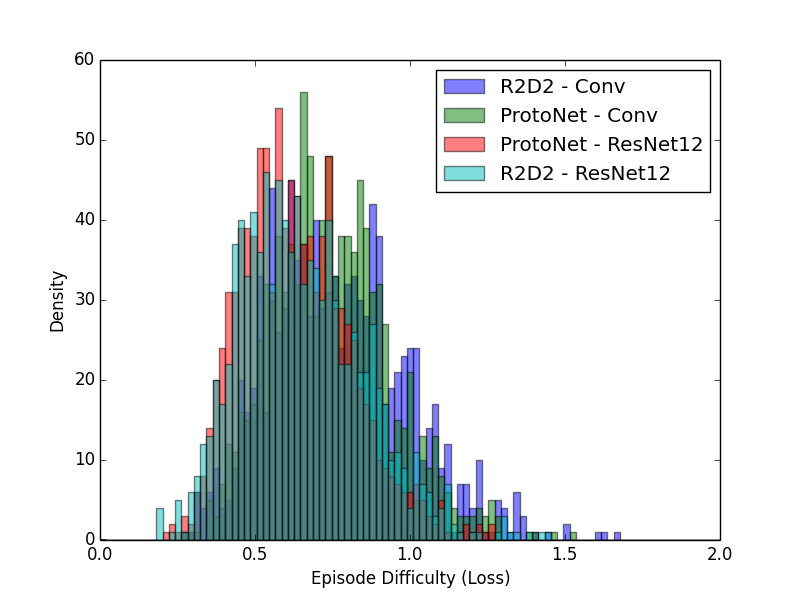}
  \caption{\label{distribution_losses} Distribution of episode hardness (difficulty) for mini-ImageNet across different meta-learners. We find that the distribution of hardness approximately follows a gamma distribution. From the density plots, we find that a significant fraction of the episodes lie in the range of medium-hard range of difficulty. }
\end{figure}
\section{Implementation Details}
\subsection{Hyperparameters}
Similar to \citep{meta_opt_net}, for the optimizer we use SGD with Nesterov momentum and weight decay. We set the momentum to be 0.9 and weight decay to be 0.0005.  During meta-training, the underlying meta-learner is trained for 60 epochs with each epoch consisting of 1000 episodes. We use 8 episodes per batch during the meta-learner update. The initial learning rate is kept at 0.1 which is then decayed using a learning rate scheduler at epoch 20, 40 and 50 as shown in \citep{meta_opt_net}. We use a 5-way classification for each of our experimental settings. During each iteration of meta-training, we use 6 query examples, while during meta-testing we use 15 query examples. Following the practice of \citep{dynamic_without_forgetting, predict_param, meta_opt_net}, we use horizontal flip, random crop, and color jitter as data augmentation techniques during meta-training. Across each of the experimental setup, the best model for meta-testing is chosen based on the best validation set performance. In general, few-shot learning can be performed in two scenarios: (i) Inductive setting where each test example is evaluated independently; (ii) Transductive setting where the few-shot learner has access to all the test examples. In all our experimental settings, we follow the inductive setting where each query example is evaluated independently. For general adversarial training method (AT), we choose sample 4 additional episodes per sampled episode. For a batch of size 8, there would be 32 additional episodes with a total of 40 episodes. 8 episodes with the highest loss are selected from this pool for optimizing the meta-learner. For our adversarial curriculum learning method (ACT), we choose the set of 8 episodes with the lowest loss for the first 30 epochs of meta-training and choose the set of 8 episodes during the last 30 epochs of meta-training. Across all our experimental settings, we use the same number of shots during meta-training and meta-testing to match the training and testing conditions.
\begin{figure*}
    \hskip-1.2cm
  \includegraphics[width=16.5cm, height=9cm]{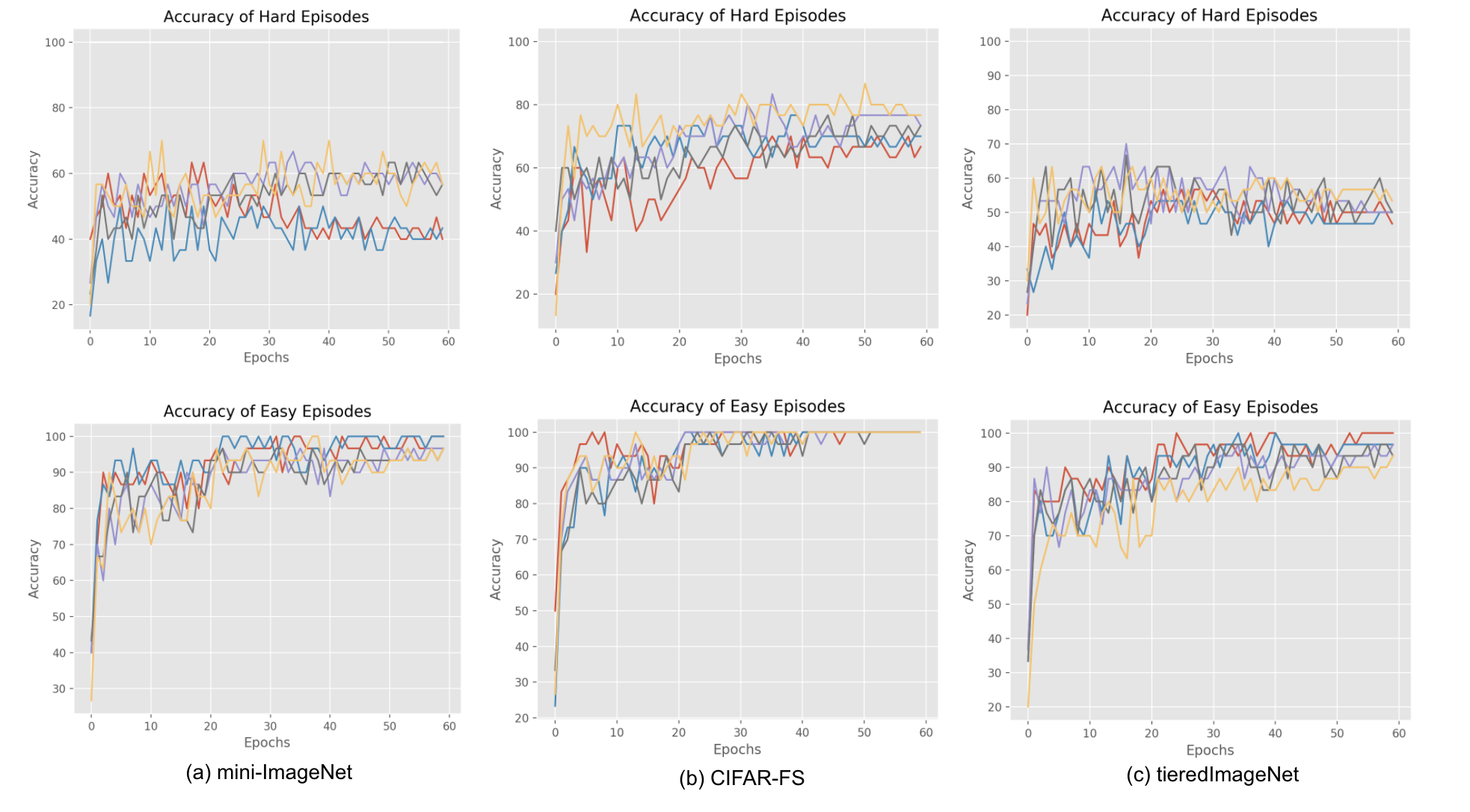}
  \vspace{-2em}
  \caption{\label{global_events_total} Accuracy of hard and easy episodes (y-axis) during the course of meta-training across different epochs (x-axis). Different colors signify different episodes. }
\end{figure*}
\subsection{Few-shot Datasets}
\textbf{mini-ImageNet . } The mini-ImageNet dataset is introduced by \citep{mini_imagenet} and is a standard few-shot classification benchmark. This dataset consists of 100 classes which are chosen from  ILSVRC-2012 \citep{imagenet1k}. The 100 classes are split into 64, 16, and 20 classes for meta-training, meta-validation and meta-testing. Each class has 600 images resulting in a total of 60000 images for the entire dataset. 

\textbf{tieredImageNet.} The tieredImageNet dataset is a larger and more challenging few-shot learning benchmark. It consists of 608 classes in total which are subclasses of ILSVRC-2012 \citep{imagenet1k}. The total number of classes are split as 351, 97 and 160 classes for meta-training, meta-validation and meta-testing respectively. These classes are selected such that there is minimum semantic similarity between the different splits, which makes this dataset challenging.

\textbf{CIFAR-FS.} The CIFAR-FS dataset \citep{R2D2} is curated from CIFAR-100 and comprises of 100 classes in total with each class consisting of 600 images. The classes are randomly split into 64, 16 and 20 for meta-training, meta-validation and meta-testing respectively. 

Amongst the three datasets, the images are of size 84x84 for mini-ImageNet and tieredImageNet, while for CIFAR-FS, the images are of size 32x32.

\section{Global Forgetting Events}
To gain further insights into global forgetting events, we select 5 hard and easy episodes from each of the few-shot datasets and track their accuracy during the course of meta-training. We use prototypical networks as the base meta-learner for our analysis. We find and validate in Fig. (\ref{global_events_total}), that our analysis on global forgetting (as shown in the main paper) holds true for more number of hard episodes. Additionally, we also choose 25 hard and easy episodes to compute the mean of their maximum accuracy reached during meta-training and the accuracy at the end of training. We find that for hard episodes on an average, the gap between the maximum accuracy reached during training and the final accuracy is large. However, for easy episodes we find this gap to be narrow. This large gap for hard episodes signifies they they undergo catastrophic forgetting during the course of meta-training. 
% \begin{figure}[H]
%     \hskip-1.2cm
%   \includegraphics[width=16.5cm, height=9cm]{composite_local_forget.png}
%   \vspace{-2em}
%   \caption{\label{global_events_total} Accuracy of hard and easy episodes (y-axis) during the course of meta-training across different epochs (x-axis). Different colors signify different episodes. }
% \end{figure}

\begin{figure}[H]
    \hskip-1.2cm
  \includegraphics[width=16.3cm, height=6cm]{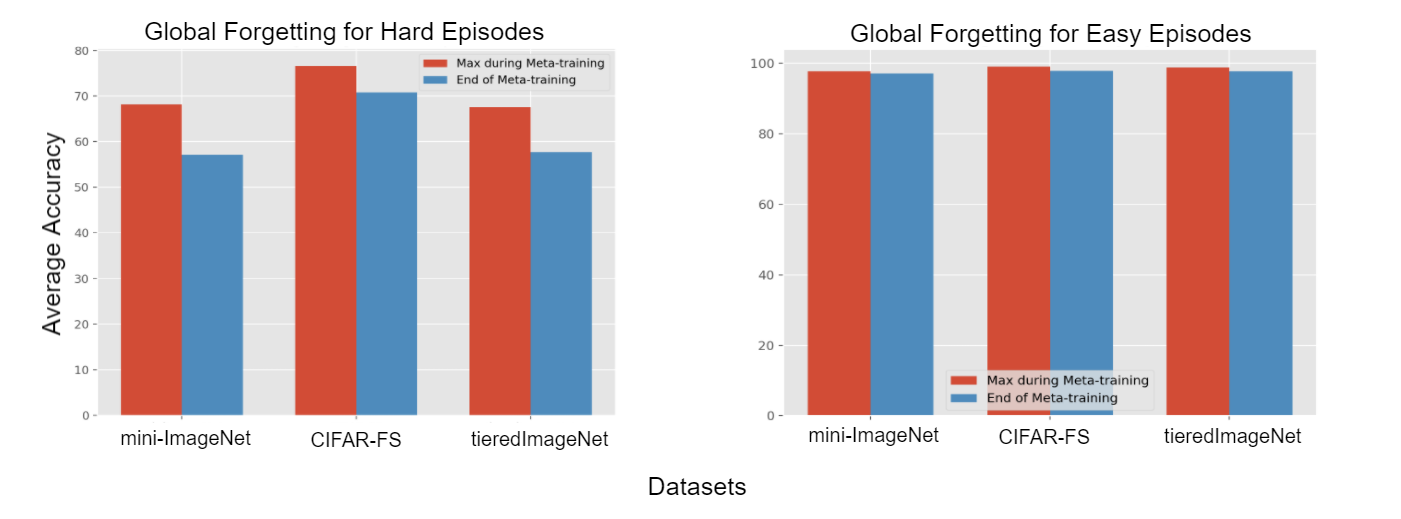}
  \vspace{-2em}
  \caption{\label{global_events_acc_new} For a set of 25 hard and easy episodes, we report the mean of the maximum accuracy reached during the course of meta-training and the accuracy at the end of meta-training. For hard episodes, we observe a substantial gap denoting global forgetting.}
\end{figure}
\section{On Transferability of Episodes}
Recent and concurrent work such as \citep{arnold2021uniform} shows that episode hardness transfer across different architectures and meta-learners. In this section, we revisit their analysis and take a closer look at the transferability of episodes across various meta-learners and architectures to obtain certain new insights. To this end, we compute the Pearson and Spearman correlations of the losses (hardness) incurred by episodes across different architectures and meta-learners for each few-shot dataset.
\begin{figure}[H]
    \hskip 1cm
  \includegraphics[width=12cm, height=10cm]{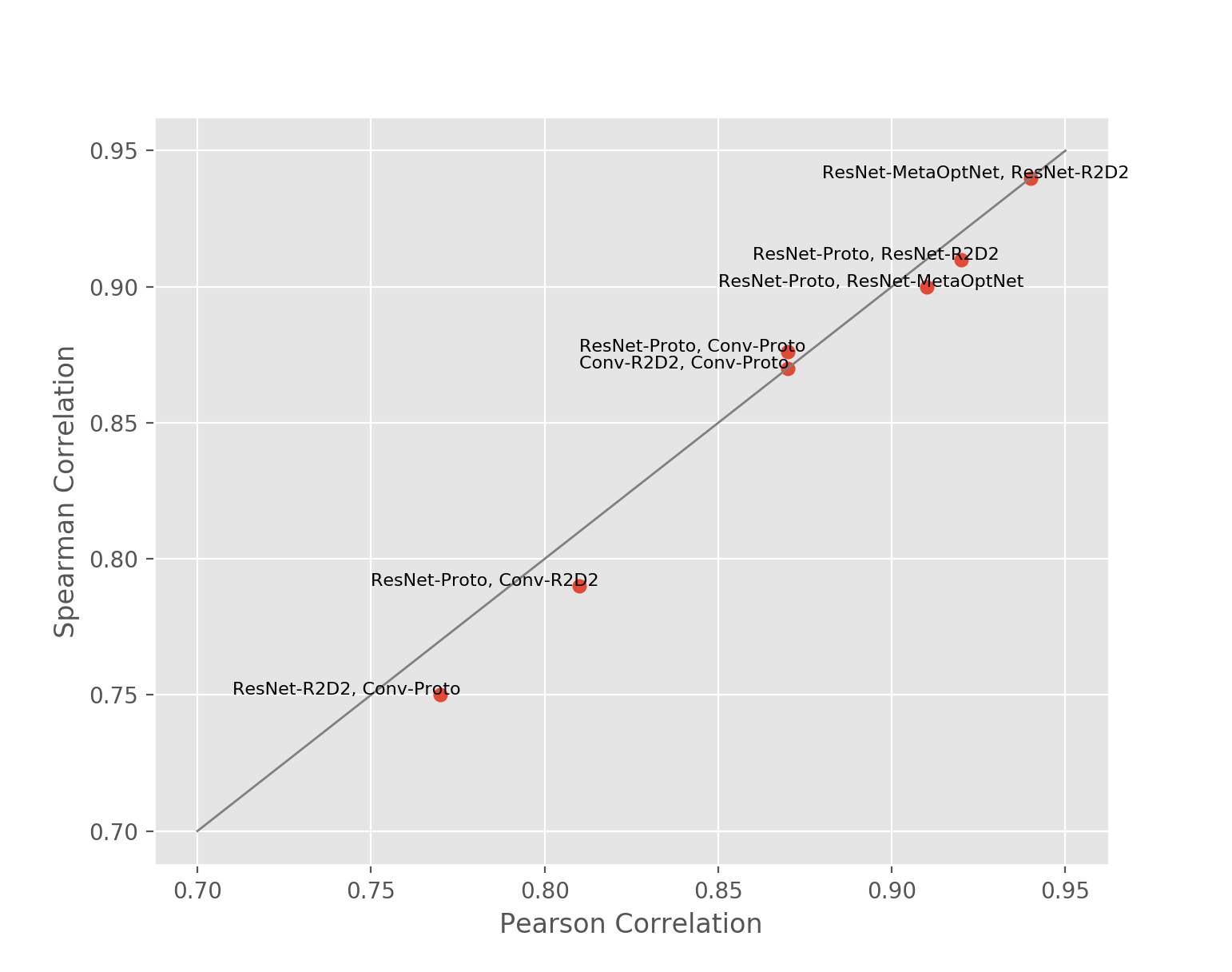}
  \caption{\label{transfer-tiered} Transferability of episode hardness for tieredImageNet.}
\end{figure}
\begin{figure}[H]
    \hskip 1cm
  \includegraphics[width=12cm, height=10cm]{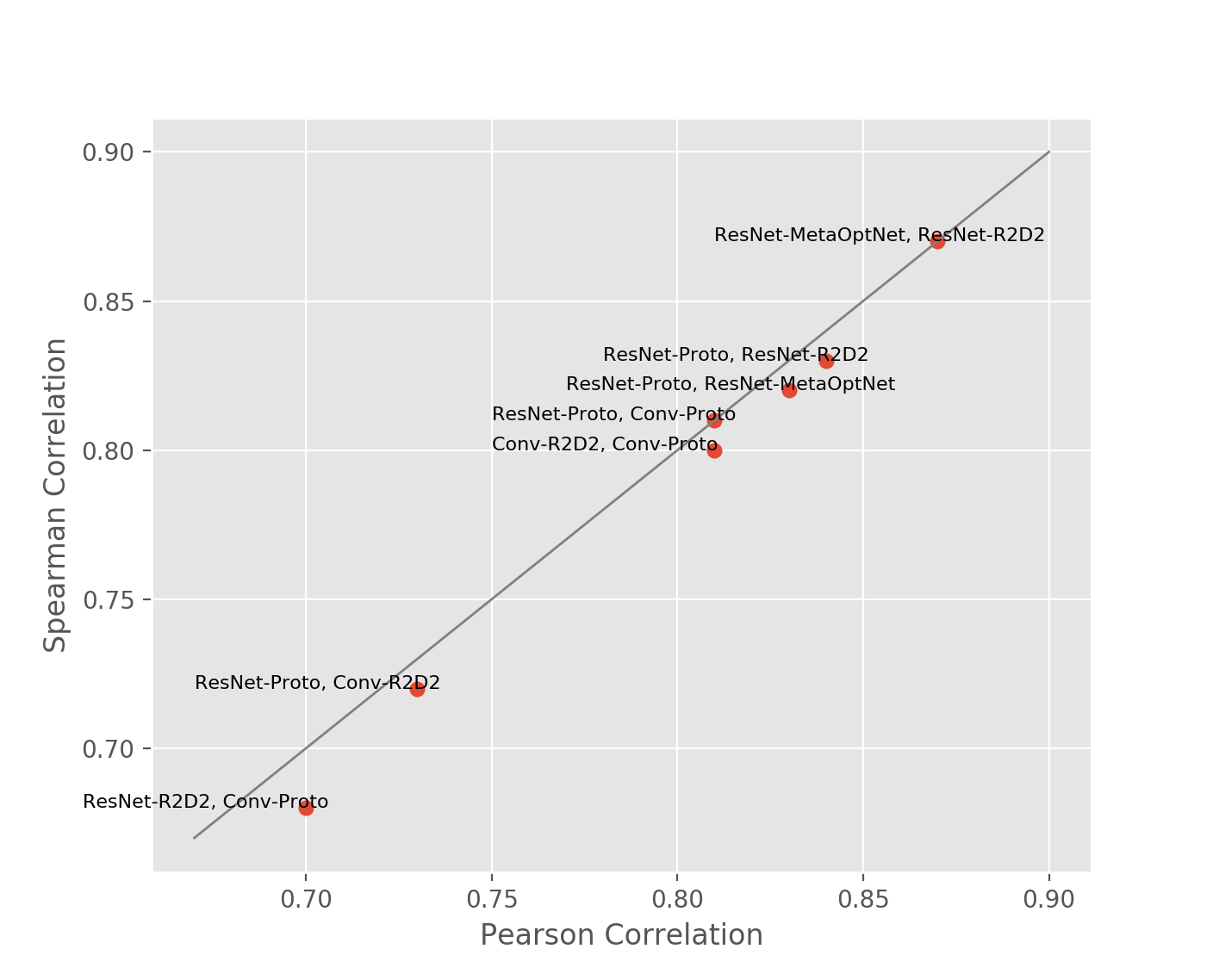}
  \caption{\label{transfer-mini} Transferability of episode hardness for mini-ImageNet.}
\end{figure}

\begin{figure}[H]
    \hskip 1cm
  \includegraphics[width=12cm, height=9.5cm]{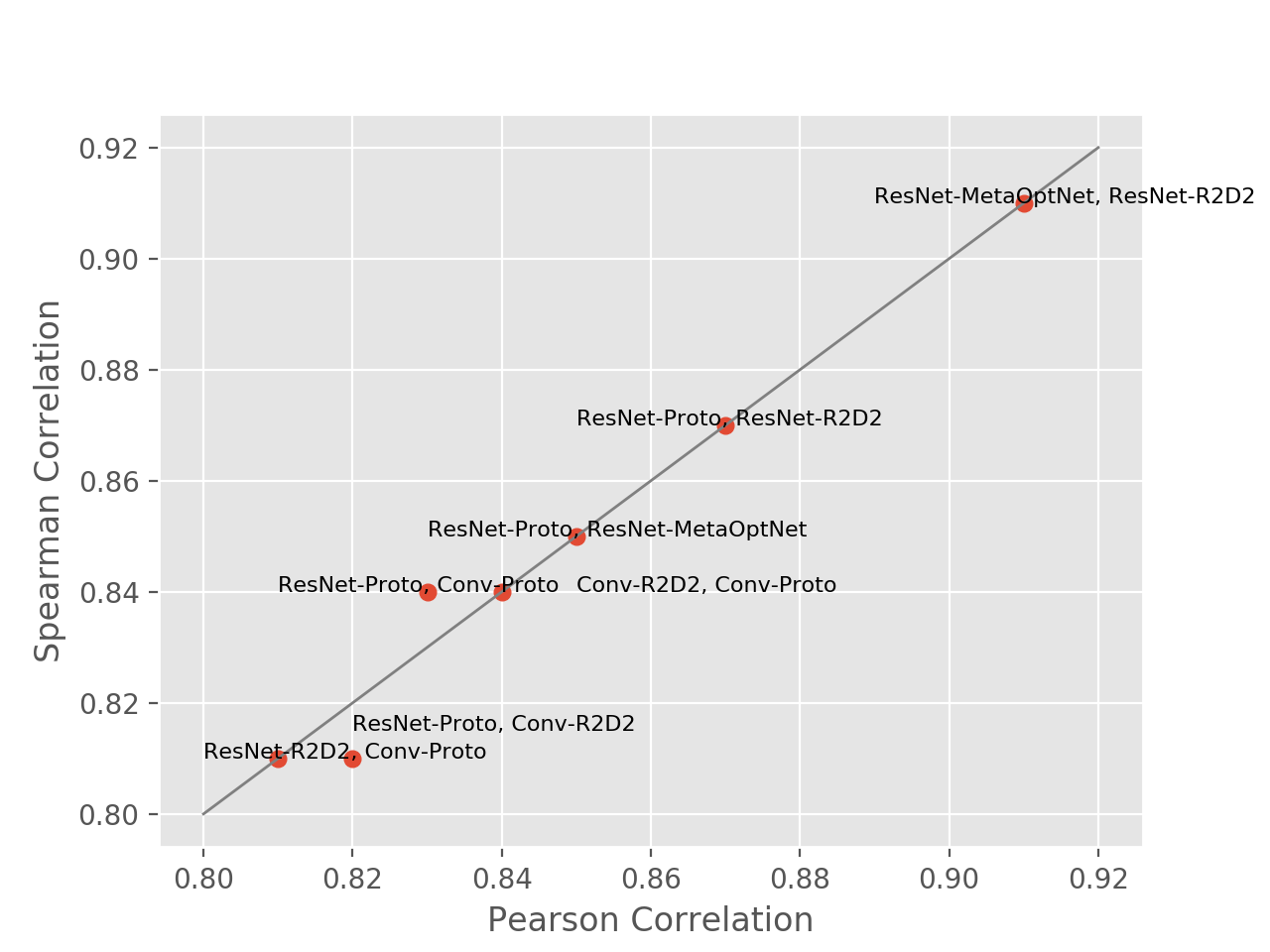}
  \caption{\label{transfer-cifar} Transferability of episode hardness for CIFAR-FS.}
\end{figure}
Across all the three few-shot datasets, we find that the order of transferability across meta-learners and architectures is similar. For example, episodes transfer the best consistently between ResNet-MetaOptNet and ResNet-R2D2 across all the datasets, while episodes transfer the worst between ResNet-R2D2 and Conv-Proto. This observation makes intuitive sense as the meta-training strategies for both R2D2 and MetaOptNet are more similar for than R2D2 and prototypical networks. Amongst the datasets, we find that transfer of episode hardness between meta-learners and architectures is much more effective for tieredImageNet than mini-ImageNet. For example, out of the 7 combinations for transferability of episodes, 3 of them have a correlation of more than 0.90 in case of tieredImageNet.

\section{More on Visual Semantics of Hard Episodes}
\begin{figure}[H]
    \hskip -2.4cm
  \includegraphics[width=19cm, height=11cm]{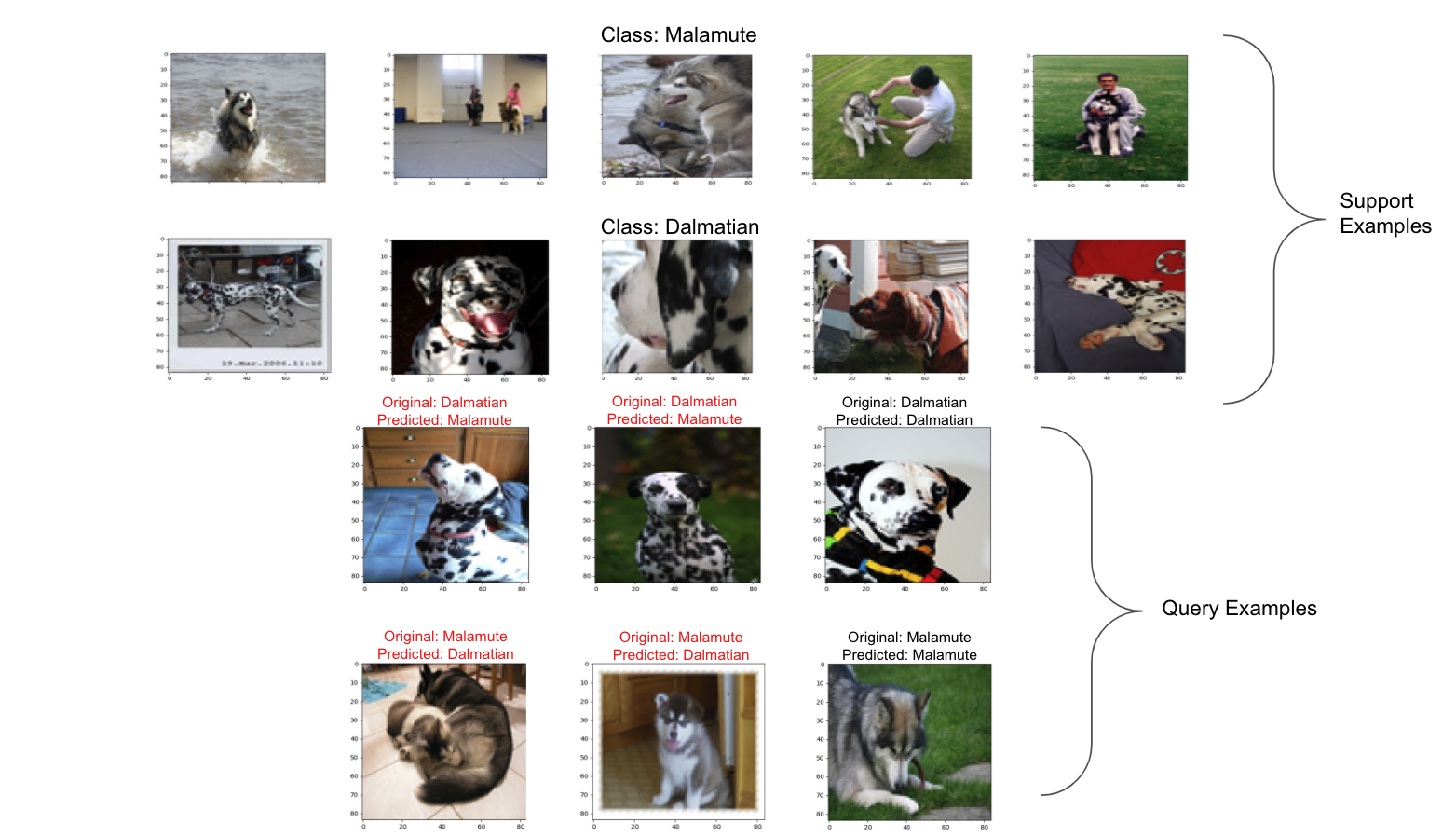}
  \caption{\label{semantics}Visual semantic properties of hard episodes: We find that if there are closely related fine-grained categories in an episode (e.g, { \it Malamute} vs. {\it Dalmatian}), the meta-learner (prototypical networks + ResNet-12) often gives a wrong prediction.}
\end{figure}
In this section, we provide more instances of hard episodes. In particular, we find that episodes also become hard when there are closely related fine-grained categories in the episode. For example, in an episode consisting of dogs from the classes {\it Malamute} and {\it Dalmatian}, we notice that the underlying meta-learner often gets confused between the two fine-grained categories leading to erroneous predictions on the query examples. Presence of similar fine-grained categories in episodes is therefore one failure mode of existing meta-learning methods.

\end{document}